\documentclass[twoside]{article}

\usepackage[accepted]{aistats2025_tweak}

\usepackage{multirow}
\usepackage{graphicx}

\usepackage[round]{natbib}

\usepackage{amsfonts,amsmath,amssymb,amsthm}
\newcommand{\cvec}{\mathbf{c}}

\newcommand{\xvec}{\mathbf{x}}

\newcommand{\Dcal}{\mathcal{D}}
\newcommand{\Ncal}{\mathcal{N}}
\newcommand{\Lcal}{\mathcal{L}}
\DeclareMathOperator*{\argmax}{arg\,max}

\usepackage{mathtools}
\newcommand{\defeq}{\vcentcolon=}

\usepackage[nameinlink]{cleveref}
\usepackage{pdflscape}

\begin{document}
%

%

\twocolumn[

\aistatstitle{Likelihood approximations via Gaussian approximate inference}

\aistatsauthor{Thang D.~Bui}

\aistatsaddress{Australian National University, Australia} ]

\begin{abstract}

Non-Gaussian likelihoods are essential for modelling complex real-world observations but pose significant computational challenges in learning and inference. Even with Gaussian priors, non-Gaussian likelihoods often lead to analytically intractable posteriors, necessitating approximation methods. To this end, we propose efficient schemes to approximate the effects of non-Gaussian likelihoods by Gaussian densities based on variational inference and moment matching in transformed bases. These enable efficient inference strategies originally designed for models with a Gaussian likelihood to be deployed. Our empirical results demonstrate that the proposed matching strategies attain good approximation quality for binary and multiclass classification in large-scale point-estimate and distributional inferential settings. In challenging streaming problems, the proposed methods outperform all existing likelihood approximations and approximate inference methods in the exact models. As a by-product, we show that the proposed approximate log-likelihoods are a superior alternative to least-squares on raw labels for neural network classification.
\end{abstract}

\section{Introduction}

Non-Gaussian likelihoods play a crucial role in modelling complex real-world observations such as categories or counts, yet they present significant computational challenges in learning and inference.
Even for a linear model with a Gaussian prior over the model parameters, non-Gaussian likelihoods typically result in analytically intractable posteriors.
Addressing this inference issue is an active research area, with a diverse set of methods ranging from generic to model-tailored, and approximate to asymptotically exact inference strategies \citep[see e.g.][]{ghahramani2015probabilistic,blei2017variational,gelman1995bayesian}.

In this paper, we investigate an unconventional approach: {\it efficiently} and {\it accurately} approximating the impact of non-Gaussian likelihoods with Gaussian distributions.
If successful, the resulting approximate model would be amenable to well-established inference strategies designed for models with Gaussian likelihoods.
One naive solution to this problem involves treating raw observed outputs as continuous, real-valued data and directly using a Gaussian likelihood. For instance, multiclass classification problems can be viewed as single-output regression on numerical labels \citep{wilson2016deep} or multi-output regression on one-hot label vectors \citep{demirkaya2020exploring,hui2020evaluation}. Whilst these approaches can achieve comparable test accuracies to those produced by the original likelihood in (regularised) maximum likelihood settings, transforming the predictions from such models into valid predictive distributions in the original observation domain remains challenging, often necessitating an expensive post-hoc calibration step \citep{platt1999probabilistic,niculescu2005predicting}.

Approximating a non-Gaussian likelihood by a Gaussian, equivalently using a least-squared loss on non-Gaussian observations, offers benefits far beyond the predictive performance in the point-estimate settings or the ease of theoretical analyses \citep{han2021neural}. In probabilistic supervised learning settings such as Gaussian process classification, a Gaussian likelihood enables closed-form and tractable posterior and marginal likelihood for hyperparameter optimisation, sidestepping potentially cumbersome approximate inference \citep[][sec 6.5]{kuss2006gaussian,williams2006gaussian}. In the large data regime, it also removes the need to parameterise an approximate posterior in sparse inducing-point variational inference \citep{titsias09a,milios2018dirichlet}. In online learning settings with streaming data, a Gaussian likelihood allows efficient, closed-form updates for Gaussian process models \citep{csato2002sparse,BuiNguTur17} or last layers in neural networks \citep{titsias2019functional,titsias2023kalman}.

Motivated by these potential benefits, we first revisit the recently proposed likelihood approximation scheme, Laplace Matching \citep{hobbhahn2022laplacematching}, and propose its relatives based on variational inference and moment matching in transformed bases for general non-Gaussian likelihoods (\cref{sec:methods}). In a suite of benchmark binary and multiclass classification problems, the proposed approaches achieve good approximation quality in both point-estimate (\cref{sec:exp_net}) and distributional inferential settings (\cref{sec:exp_gpc}), where the Laplace variant often underperforms. Our empirical results also demonstrate the superior performance-efficiency frontiers of the proposed likelihood approximations in challenging practical streaming settings (\cref{sec:exp_streaming}). 


\section{Background}
We first concisely describe the necessary background, including likelihoods in supervised learning and some properties related to the Dirichlet distributions. For clarity, we will focus on the parametric treatment of supervised learning models, but the proposed approximation can be applied to nonparametric and also latent variables models.
\subsection{Supervised learning}
Suppose we have a training set comprising $N$ $D$-dimensional input vectors $\{\xvec_n\}^N_
{n=1}$ and corresponding scalar observations $\{y_n\}_{n=1}^{N}$. For regression, $y_n$ is typically continuous and real-valued. For K-class classification, $y_n \in \{1, 2, \dots, K\}$ and could be equivalently represented by one-hot vectors, $\cvec_n = [c_{n1}, \dots, c_{nK}]$, where $c_{nk} = 1$ if $y_n = k$ and $0$ otherwise. We model the latent function by placing a prior over the parameters $\theta$, $p(\theta)$, or directly over $f$, $p(f_\theta)$. To complete the model specification, we specify the likelihood of the parameters, or the function given an observation, $p(y_n | f_\theta(\xvec_n))$. Some popular forms for the likelihood are: for regression, $p(y_n | \theta, \xvec_n) = \Ncal(y_n;  f_\theta(\xvec_n), \sigma_y^2)$, where $\sigma_y^2$ is the observation noise variance; for binary classification, $p(y_n| \theta, \xvec_n) = \phi(f_\theta(\xvec_n))^{y_n}(1- \phi(f_\theta(\xvec_n)))^{1 - y_n}$ where $\phi$ is the logistic sigmoid; for multiclass classification $p(y_n| \theta, \xvec_n) = \prod_k \phi(f_{\theta, 1:K}(\xvec_n))_k^{c_{nk}}$ where $\phi$ is the softmax function. The Gaussian likelihood on raw labels can be used for classification and is often termed label regression or least-squares for classification. While not grounded by a principled probabilistic description of the categorical outputs, it is known to achieve good prediction accuracy for neural networks and Gaussian processes \citep{kuss2006gaussian,demirkaya2020exploring,hui2020evaluation}.

The typical learning task involves finding $\theta$ by optimising the regularised maximum likelihood objective to get a point estimate, $\theta_p = \argmax_\theta{\Lcal(\theta)}$, where
\begin{align}
   \Lcal(\theta) = \log p(\theta) + \sum_n \log p(y_n | \theta, \xvec_n),\label{eqn:map}
\end{align}
or computing the posterior belief about $\theta$,
\begin{align}
    p(\theta | \Dcal \coloneq \{\xvec_n, y_n\}_n) \propto p(\theta) \prod_n p(y_n | \theta, \xvec_n). \label{eqn:bayes}
\end{align}
The latter approach is analytically intractable for most models and thus requires approximations such as variational inference, expectation propagation and Laplace approximation, or asymptotically exact schemes like MCMC. The resulting point-estimate or (approximate) posterior belief is subsequently used for making predictions at new test points, $p(y^* | \theta_p, \xvec^* )$ or $p(y^* | \xvec^*, \Dcal) = \int_\theta p(y^* | \theta, \xvec^* ) p(\theta | \Dcal)$.

\subsection{Dirichlet distribution, construction and usage as priors}
The Dirichlet distribution is central to the proposed likelihood approximation to the softmax likelihood. We will therefore state its definition and relevant properties. It is a multivariate probability distribution, defined on the $K-1$-dimensional simplex $\Delta^{K-1} \defeq {\boldsymbol{\pi} \in \mathbb{R}^K : \pi_k \geq 0, \sum_k \pi_k = 1}$ and parameterised by concentration parameters $\boldsymbol{\alpha} = [\alpha_1, \dots, \alpha_K]$. The probability density function (pdf) of a Dirichlet random variable $\boldsymbol{\pi} \sim \mathrm{Dir}(\boldsymbol{\alpha})$ is given by $p_{\boldsymbol{\pi}}(\boldsymbol{\pi}) = B^{-1}(\boldsymbol{\alpha}) \prod_k \pi_k^{\alpha_k - 1}$
where $B(\boldsymbol{\alpha})$ is the multivariate Beta function.

An interesting fact is the Dirichlet distribution can be constructed using independent Gamma variables. Let $\omega_1, \ldots, \omega_K$ be independent Gamma random variables with $\omega_k \sim \mathrm{Gamma}(\alpha_k, 1)$. Define $\pi_k = \omega_k / \sum_{j=1}^K \omega_j$ for $k = 1, \ldots, K$. Then, $\boldsymbol{\pi} = (\pi_1, \dots, \pi_K)$ follows a Dirichlet distribution with parameter $\boldsymbol{\alpha}$, i.e., $\boldsymbol{\pi} \sim \mathrm{Dir}(\boldsymbol{\alpha})$ \citep[see e.g.][]{frigyik2010introduction}.

\begin{figure*}[!ht]
    \vspace{-12pt}
    \centering
    \includegraphics[width=\linewidth]{{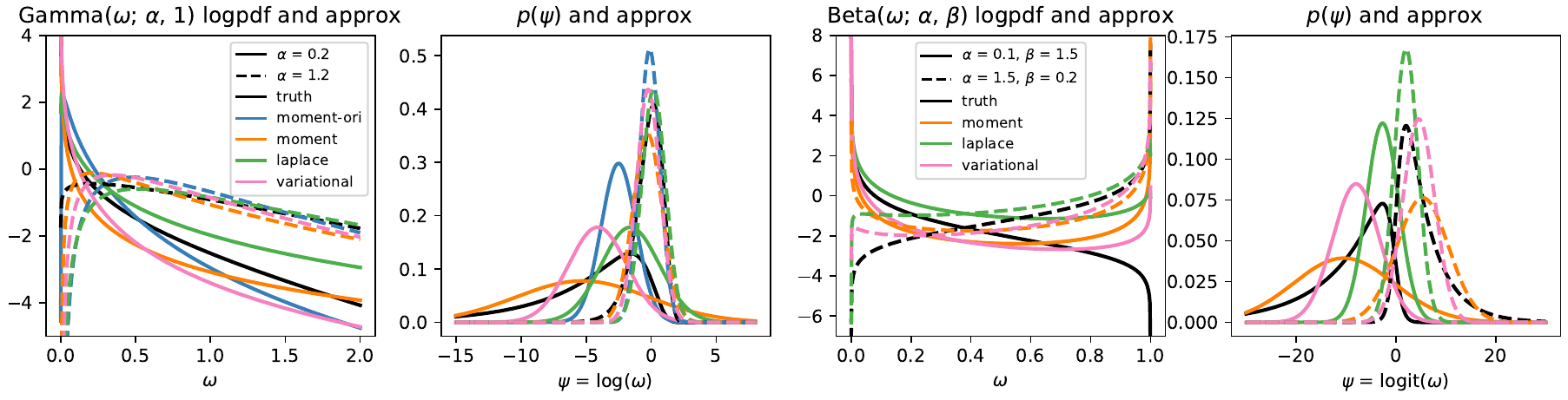}}
    \vspace{-22pt}
    \caption{Approximations of Gamma densities [first two plots, \cref{sec:gamma_multiclass}] and Beta densities [last two plots, \cref{sec:beta_binary}]. The methods included are log-Normal moment matching \citep[][moment-ori]{milios2018dirichlet}, Laplace approximation \citep[][laplace]{hobbhahn2022laplacematching}, and Gaussian moment matching (moment) and variational matching (variational) in transformed bases. The first and third columns show the log-pdfs and their approximations. The second and fourth plots show the corresponding log/logit-transformed densities and their Gaussian approximations. Best viewed in colour.}
    \label{fig:gamma_beta_approx_viz}
    \vspace{-12pt}
\end{figure*}

The Dirichlet distribution is conjugate to the multinomial likelihood, making it a convenient prior choice for modelling categorical data. Specifically, the likelihood of observing counts $\boldsymbol{c} = (c_1, \ldots, c_K)$ given the probability vector $\boldsymbol{\pi}$ is $p(\boldsymbol{c} | \boldsymbol{\pi}) \propto \prod_{k=1}^K \pi_k^{c_k}$.
When a Dirichlet prior $\boldsymbol{\pi} \sim \mathrm{Dir}(\boldsymbol{\alpha})$ is placed on $\boldsymbol{\pi}$, the posterior distribution of $\boldsymbol{\pi}$ given the observed counts $\boldsymbol{c}$ is also Dirichlet: $p(\boldsymbol{\pi} | \boldsymbol{c}, \boldsymbol{\alpha}) \propto p(\boldsymbol{c} | \boldsymbol{\pi}) p(\boldsymbol{\pi} | \boldsymbol{\alpha}) \propto \mathrm{Dir}(\boldsymbol{\pi}; \boldsymbol{\alpha} + \boldsymbol{c})$.

\section{Methodology}
\label{sec:methods}
The goal of our proposed approach is to approximate the effect of the complex non-Gaussian likelihoods on the posterior in \cref{eqn:bayes} or the regularised log-likelihood in \cref{eqn:map} by a Gaussian density. While this might sound like expectation propagation or variational message passing to familiar readers, the likelihood approximations here are constructed only once and do not require further iterative tuning. We now discuss a variational inference strategy to approximate common exponential family distributions in a transformed basis and illustrate how the resulting approximations can be used for binary or multiclass classification. Our approach builds on the Laplace approximation of \cite{hobbhahn2022laplacematching}.

\subsection{Transformed exponential family densities and approximation in transformed basis}
Let $\omega \in \Omega $ be a $d-$dimensional random variable in the exponential family 
and $g: \Omega \rightarrow \mathbb{R}^d, \psi = g(\omega)$ be an invertible, differentiable transformation of $\omega$ with inverse $\omega = g^{-1}(\psi)$. The pdf of $\psi$ can be obtained using the change of variables formula:
\begin{align*}
p_{\psi}(\psi) = p_{\omega}(g^{-1}(\psi)) \left| \frac{\partial g^{-1}(\psi)}{\partial \psi}  \right|.
\end{align*}
The transformation $g$ is chosen such that the support of $\psi$ is $\mathbb{R}^d$ and that the density in the new basis, $p_{\psi}(\psi)$, has a (usually skewed) bell shape and thus can be well-approximated by a Gaussian distribution. \cite{hobbhahn2022laplacematching} use Laplace's method to find a Gaussian whose mean and covariance match the mode and negative Hessian of $\log p_{\psi}(\psi)$, respectively. We find that this approach is inferior in many practical use cases, motivating alternative approximation strategies based on Gaussian approximate inference.

\begin{figure*}[ht!]
    \centering
    \vspace{-10pt}
    \includegraphics[width=1.0\linewidth]{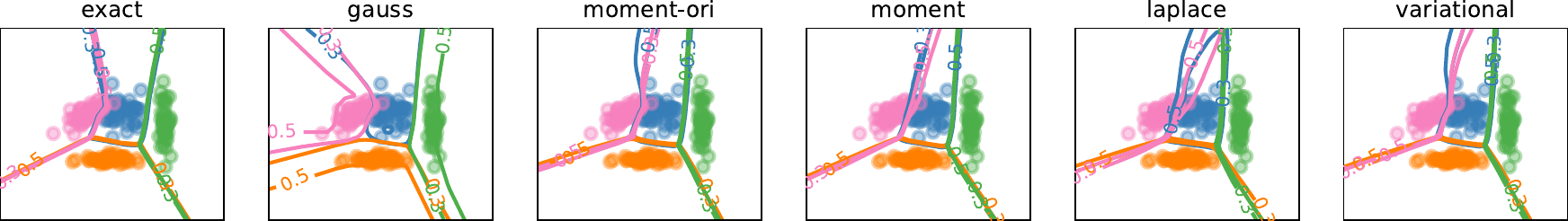}
    \vspace{-18pt}
    \caption{Neural networks classifiers on a toy four-class dataset trained by softmax cross-entropy [exact], one-hot label regression [gauss], and mean-squared errors using log-normal moment matching in the standard basis [moment-ori], and Gaussian moment matching [moment], Laplace matching [laplace] and variational matching [variational] in the log basis. Each network has two hidden layers, each with 64 rectified linear units.}
    \label{fig:toyclass}
    \vspace{-12pt}
\end{figure*}

\paragraph{Variational matching} Our aim is to approximate the true density $p_{\psi}(\psi)$ with a simpler, Gaussian distribution $q(\psi) = \mathcal{N}(\psi; \mu, \Sigma)$. The mean and covariance can be found analytically or numerically by minimising the following variational objective,
\begin{align*}
    \mathcal{F}(q) = \text{KL}(q(\psi) || p_\psi(\psi)) = \mathbb{E}_{q(\psi)}[\log q(\psi) - \log p_\psi(\psi)].
\end{align*}

\paragraph{Moment matching} Alternatively, we can minimise $\text{KL}(p_\psi(\psi) || q(\psi))$, which is to match the first and second moments of $q(\psi)$ to that of $p_\psi(\psi)$. These moments can be found analytically for many classes of base densities and transformations, using direct calculation or the moment-generating function.

We will now work through the above approximations for typical densities and transformations.

\subsection{Gamma distribution and $\log$ basis, and multiclass classification\label{sec:gamma_multiclass}}
If $\omega \sim \mathrm{Gamma}(\alpha, \beta)$ and $\psi = g(\omega) = \log(\omega)$, then $\omega = g^{-1}(\psi) = \exp(\psi)$, $\frac{\partial g^{-1}(\psi)}{\partial \psi} = \exp(\psi)$, and thus $p_{\psi}(\psi) = \mathrm{Gamma}(\exp(\psi); \alpha, \beta) \exp(\psi)$. 
\paragraph{Laplace matching} The mode and second derivative of $\log p_{\psi}(\psi) = c - \beta \exp(\psi) + \alpha \psi$ are available in closed-form, leading to $q(\psi) = \Ncal(\psi; \log(\alpha/\beta), \alpha^{-1})$ \citep{hobbhahn2022laplacematching}.
\paragraph{Variational matching} For $q(\psi) = \Ncal(\psi; \mu, \sigma^2)$, the variational objective is $\mathcal{F}(q) = c - 0.5 \log(\sigma^2) - \alpha \mu + \beta \exp(\mu + 0.5 \sigma^2)$. Setting the gradients wrt.~$\mu$ and $\sigma^2$ to 0 gives $q(\psi) = \Ncal(\psi; \log(\alpha/\beta) - 0.5 \alpha^{-1}, \alpha^{-1})$
\paragraph{Moment matching} We first compute the mean of $\psi$: $\mathbb{E}_\psi[\psi] = \mathbb{E}_\omega[\log(\omega)] = \Psi(\alpha) - \log(\beta)$ where $\Psi$ is the digamma function. Similarly, the variance $\mathbb{V}_\psi[\psi] = \Psi^{1}(\alpha)$ where $\Psi^1$ is the trigamma function. Therefore, $q(\psi) = \Ncal(\psi; \Psi(\alpha) - \log(\beta), \Psi^1(\alpha))$.

Readers may have noted that Gaussian approximations in the $\log$ basis correspond to using a log-normal distribution to approximate the original Gamma distribution. In fact, \cite{milios2018dirichlet} note a moment-matched log-normal in the original basis can be obtained, corresponding to the following distribution in the $\log$ basis: $q(\psi) = \Ncal(\psi; \log(\alpha/\beta) - 0.5 \sigma^{2}, \sigma^2)$ where $\sigma^2 = \log (1 + \beta / \alpha)$. We visualise the above approximations, including moment matching in standard basis of \cite{milios2018dirichlet} [\textit{moment-ori}], Laplace matching of \cite{hobbhahn2022laplacematching} [\textit{laplace}], moment matching [\textit{moment}] and variational matching [\textit{variational}] in \cref{fig:gamma_beta_approx_viz}. The Gaussian approximations are similar and qualitatively more accurate for large $\alpha$'s. For small $\alpha$'s, \textit{laplace} places more mass in the zero-density region, \textit{moment} has a large variance, while \textit{moment-ori} and \textit{variational} are visually reasonable.

\paragraph{Softmax likelihood\footnote{Strictly speaking, `softmax' is a function but we are overloading it here to mean a distribution over categories whose probabilities are the softmax outputs.} approximations} We can now apply our approximation methods to $K$-class classification problems. The presentation here is based on \citep{milios2018dirichlet} and \citep{wang2023gpp} but we provide a clearer, step-by-step exposition. The high-level intuition is that the softmax likelihood measures how the $K$-dimensional logits, the outputs of the underlying model, can explain each observation. We first treat the exponentiated logits for each data point as random variables and obtain an auxiliary posterior for them. We will then log transform an equivalent representation of this posterior and use the above approximations to derive auxiliary Gaussian distributions over the logit variables. These distributions can subsequently be used as an alternative to the original softmax likelihood.

In detail, we first treat each observation as a count vector $\mathbf{c} = (c_1, ..., c_K)$, where $c_k = 1$ if $y = k$ and 0 otherwise. Using a Dirichlet prior $\mathrm{Dir}(\boldsymbol{\alpha}_{\epsilon})$ on the class probabilities $\boldsymbol{\pi}$ leads to the posterior in the same family, $\mathrm{Dir}(\boldsymbol{\alpha}_{\epsilon} + \mathbf{c})$. We can then utilise the construction using $K$ independent Gamma variables where $\omega_k \sim \mathrm{Gamma}(\alpha_{\epsilon,k} + c_k, 1)$, and apply a log-transformation, $\psi_k = \log(\omega_k)$, and utilise the above Gaussian approximations in this new basis. While the described path may seem winding, its logic becomes clearer when we consider the reverse process. Given the $\{\psi_k\}_{k=1}^K$ variables, we can get the Dirichlet's class probability by first exponentiating them and renormalise, $\pi_k = \exp(\psi_k) / \sum_j \exp(\psi_j)$, which is exactly the softmax warping. That is, the equivalent `exact' softmax likelihood can be approximated by $K$ `soft' Gaussian likelihoods in the logit space. The approximation quality is partly controlled by the hyperparameters $\boldsymbol{\alpha}_{\epsilon}$: smaller $\alpha_{\epsilon,k}$'s mean the Dirichlet posterior is sharper and thus closer to the original softmax likelihood but this is the regime where Gaussian approximations to the transformed Gamma density often struggle (see \cref{fig:gamma_beta_approx_viz}).

The Gaussian approximations above can be obtained with minimal computational overheads and crucially allow viewing the classification task as Gaussian regression.\footnote{For example, in gpytorch, the moment matching of \cite{milios2018dirichlet} is straightforwardly implemented as Gaussian likelihood with fixed heteroscedastic noise.} Specifically, let $\psi_{nk} = f_{\theta, k}(\xvec_n)$ be the $k$-th logit parameterised by parameters $\theta$ (or by function $f$) and a Gaussian approximation $q(\psi_{nk}) = \Ncal(\psi_{nk}; m_{n,k}, v_{n,k})$. Learning proceeds by optimising an approximation to $\Lcal(\theta)$,
\begin{align}
   \Lcal(\theta) \approx \log p(\theta) + \sum_{n,k} \log \Ncal (f_{\theta, k}(\xvec_n); m_{n,k}, v_{n,k}),\label{eqn:approx_map}
\end{align}
or computing the approximate posterior,
\begin{align}
    p(\theta | \Dcal) \approx q(\theta)  \propto p(\theta) \prod_{n,k} \Ncal(f_{\theta, k}(\xvec_n); m_{n,k}, v_{n,k}). \label{eqn:approx_bayes}
\end{align}
This paves the way for closed-form inference for a large class of models including Bayesian multinomial logistic regression, Gaussian process classification, and Bayesian last layers in neural network classification. For prediction on new data points, we can either compute $\psi$ based on the point estimate of $\theta$ in \cref{eqn:approx_map} or sample $\psi$ based on $q(\theta)$ in \cref{eqn:approx_bayes}, and warp them through the softmax function to obtain the class probabilities. \Cref{fig:toyclass} illustrates the predictions by neural network classifiers trained on a toy four-class dataset using either the cross-entropy loss corresponding to the softmax likelihood (exact) or mean squared errors corresponding to the Gaussian likelihoods whose means are the one-hot labels (gauss), and whose means and variances are provided by moment matching in standard basis \citep[][moment-ori]{milios2018dirichlet}, and variational matching (variational), moment matching (moment) and Laplace matching \citep[][laplace]{hobbhahn2022laplacematching} in the log basis.

\subsection{Beta distribution, $\mathrm{logit}$ basis, and binary classification\label{sec:beta_binary}}
If $\omega \sim \mathrm{Beta}(\alpha, \beta)$ and $\psi = g(\omega) = \mathrm{logit}(\omega) = \log(\frac{\omega}{1-\omega})$, then $\omega = g^{-1}(\psi) = \phi(\psi)$, $\frac{\partial g^{-1}(\psi)}{\partial \psi} = \phi(\psi)(1-\phi(\psi))$, and $p_{\psi}(\psi) = \mathrm{Beta}(\phi(\psi); \alpha, \beta) \phi(\psi)(1-\phi(\psi))$, with $\phi$ is the logistic sigmoid function.

\paragraph{Laplace matching}
As derived in \citep{hobbhahn2022laplacematching}, the Laplace matching approximation is:
$q(\psi) = \mathcal{N}(\psi; \log(\alpha / \beta), (\alpha + \beta) / (\alpha \beta))$.

\paragraph{Variational matching}
For $q(\psi) = \mathcal{N}(\psi; \mu, \sigma^2)$, the variational objective is:
$\mathcal{F}(q) = c - 0.5 \log(\sigma^2) + \beta \mu + (\alpha + \beta) \mathbb{E}_q[\log(1 + \exp(-\psi))]$.
This does not have a closed-form solution but we can use numerical optimisation and the reparameterisation trick \citep{rezende14,kingma2013auto} to estimate the variational parameters $\mu$ and $\sigma^2$.

\paragraph{Moment matching}
The derivation of the mean and variance of $\psi$ is more involved but gives close-form solutions, yielding  $q(\psi) = \Ncal(\psi; \Psi(\alpha) - \Psi(\beta), \Psi^1(\alpha) + \Psi^1(\beta))$. We provide the full derivation in the appendix and a visual comparison in \cref{fig:gamma_beta_approx_viz}.

\paragraph{Logistic likelihood approximations} We can apply the same recipe done for multiclass classification to the binary setting. We treat each observation as a Bernoulli trial with $y \sim \mathrm{Bernoulli}(\omega)$, where $\omega$ is the probability of the positive class. We use a Beta prior $\mathrm{Beta}(\alpha_\epsilon, \beta_\epsilon)$ on $\omega$, which leads to a Beta posterior $\mathrm{Beta}(\alpha_\epsilon + y, \beta_\epsilon + 1 - y)$.
We then apply the logit transformation $\psi = \mathrm{logit}(\omega) = \log(\omega/(1-\omega))$ and use the Gaussian approximations derived above in this logit space. The resulting approximations can be used as alternatives to the original logistic sigmoid likelihood. Specifically, let $\psi_n = f_\theta(\mathbf{x}_n)$ be the logit output of our model for the $n$-th data point. The true likelihood $p(y_n | \psi_n) = \phi(\psi_n)^{y_n}(1-\phi(\psi_n))^{1-y_n}$ can then be approximated by a Gaussian likelihood $\mathcal{N}(\psi_n; m_n, v_n)$, where $m_n$ and $v_n$ are determined by the chosen approximation method (Laplace, variational, or moment matching).
The learning and inference procedures are analogous to the multiclass case, turning models such as Bayesian logistic regression and Gaussian process binary classification to simple Gaussian linear/process regression.
The choice of $\alpha_\epsilon$ and $\beta_\epsilon$ affects the approximation quality, with smaller values leading to sharper posteriors that are closer to the original logistic likelihood but more challenging to approximate (see \cref{fig:gamma_beta_approx_viz}).

\vspace{-10pt}
\section{Experiments}
\vspace{-5pt}

To evaluate the effectiveness of our proposed Gaussian likelihood approximations, we conducted a series of experiments addressing the following key questions: (i) how accurate are the Gaussian approximations across different classification tasks and model architectures? and (ii) how do the approximations perform in much more challenging streaming data settings?. The implementation is built on standard pytorch libraries and the experiments were run on multiple V100 GPUs. Full results and additional experiments are included in the supplementary material. The hyperparameters $\alpha_\epsilon$ in \cref{sec:gamma_multiclass} and $\alpha_\epsilon$ and $\beta_\epsilon$ in \cref{sec:beta_binary} were chosen based on the log-likelihood on the train set, as suggested by \cite{milios2018dirichlet}. We can, however, sidestep this process via variational inference, as presented in the appendix.

\subsection{Approximation quality on standard neural-network classification benchmarks}
\label{sec:exp_net}
\begin{table*}[!ht]
\small
\centering
\vspace{-20pt}
\caption{Predictive performance of VGG-11 and ResNet-18 on CIFAR-10, trained with the multiclass cross-entropy loss and various Gaussian approximations\label{table:cifar-10}. The results are means and standard errors over 10 runs.}
\begin{tabular}{ccccccc}
\hline
Net & Method & Accuracy [\%, $\uparrow$] & NLL [nat, $\downarrow$] & ECE [$\downarrow$] &  CIFAR-100 AUC [$\uparrow$] &  SVHN AUC [$\uparrow$] \\ \hline
\multirow{6}{*}{vgg11} & exact & 91.05 $\pm$ 0.05 & 0.360 $\pm$ 0.003 & 0.057 $\pm$ 0.001 & 0.848 $\pm$ 0.000 & 0.858 $\pm$ 0.003\\
& gauss & 88.62 $\pm$ 0.07 & 1.598 $\pm$ 0.000 & 0.670 $\pm$ 0.001 & 0.806 $\pm$ 0.001 & 0.835 $\pm$ 0.006\\
& moment-ori & 91.33 $\pm$ 0.04 & 0.362 $\pm$ 0.002 & 0.046 $\pm$ 0.000 & 0.801 $\pm$ 0.001 & 0.826 $\pm$ 0.005\\
& moment & 89.66 $\pm$ 0.06 & 0.457 $\pm$ 0.003 & 0.067 $\pm$ 0.001 & 0.803 $\pm$ 0.001 & 0.847 $\pm$ 0.003\\
& laplace & 89.63 $\pm$ 0.05 & 0.604 $\pm$ 0.001 & 0.254 $\pm$ 0.001 & 0.818 $\pm$ 0.001 & 0.843 $\pm$ 0.004\\
& variational & 91.22 $\pm$ 0.05 & 0.366 $\pm$ 0.002 & 0.051 $\pm$ 0.000 & 0.808 $\pm$ 0.001 & 0.837 $\pm$ 0.005\\
\hline
\multirow{6}{*}{resnet18} & exact & 93.53 $\pm$ 0.03 & 0.247 $\pm$ 0.001 & 0.040 $\pm$ 0.000 & 0.875 $\pm$ 0.001 & 0.881 $\pm$ 0.003\\
& gauss & 91.15 $\pm$ 0.04 & 1.562 $\pm$ 0.001 & 0.690 $\pm$ 0.000 & 0.795 $\pm$ 0.002 & 0.788 $\pm$ 0.012\\
& moment-ori & 93.79 $\pm$ 0.05 & 0.284 $\pm$ 0.002 & 0.034 $\pm$ 0.000 & 0.788 $\pm$ 0.002 & 0.782 $\pm$ 0.010\\
& moment & 92.35 $\pm$ 0.05 & 0.365 $\pm$ 0.003 & 0.050 $\pm$ 0.001 & 0.809 $\pm$ 0.001 & 0.850 $\pm$ 0.011\\
& laplace & 92.24 $\pm$ 0.04 & 0.535 $\pm$ 0.001 & 0.251 $\pm$ 0.000 & 0.798 $\pm$ 0.001 & 0.790 $\pm$ 0.012\\
& variational & 93.70 $\pm$ 0.05 & 0.291 $\pm$ 0.002 & 0.040 $\pm$ 0.000 & 0.796 $\pm$ 0.001 & 0.811 $\pm$ 0.012\\
\hline
\end{tabular}
\vspace{-10pt}
\end{table*}

We assessed the quality of our Gaussian approximations on standard image/text classification benchmarks using various neural network architectures. We first trained a convolutional neural network (CNN) on the MNIST dataset using the regularised maximum likelihood objective and different likelihood approximations described in \cref{sec:gamma_multiclass}: moment matching in standard basis (\textit{moment-ori}), moment matching in log basis (\textit{moment}), Laplace matching (\textit{laplace}), variational matching (\textit{variational}), and Gaussian regression on one-hot labels (\textit{gauss}). We compared these to the exact softmax likelihood (\textit{exact}), and reported test error and test loss curves in \cref{fig:mnist_best} in the appendix due to space constraints. In terms of test accuracy, all approximation methods, including the naive gauss approach, converge to values close to the exact likelihood. For the test loss, the variational and moment approximations perform closest to the exact likelihood, with the loss nearly matching the exact method. The exact cross-entropy loss suffers from slight overfitting while the `soft' Gaussian approximations can avoid this. Laplace and moment-ori methods are inferior but much better than gauss. 

We extended our evaluation to the more challenging CIFAR-10 dataset, using both VGG-11 and ResNet-18 architectures. \Cref{table:cifar-10} presents the results, including accuracy, negative log-likelihood (NLL), expected calibration error (ECE), and out-of-distribution (OOD) detection performance on CIFAR-100 and SVHN datasets, measured by the AUC when using the maximum predictive probability as the criterion to distinguish in-distribution and OOD. We note the variational and moment-ori approximations consistently achieve accuracies closest to (or even slightly better than) the exact likelihood for both architectures. The moment and laplace approximations show slightly lower accuracies but outperform the gauss baseline. For the NLL and ECE metrics, variational and moment-ori approximations again perform best, with NLL and ECE very close to the exact likelihood. The laplace and moment approximations are less accurate in terms of probabilistic predictions and calibration, but are still better than gauss. For OOD detection, all approximations generally perform similarly but suffer slightly compared to the exact likelihood.

Finally, we compared the approximations in \cref{sec:beta_binary} to the binary cross entropy loss for a binary classification task. We finetuned a pretrained DistilBERT model \citep{sanh2019distilbert} on the IMDB review sentiment dataset \cite{maas-etal-2011-learning} and summarised the performance in \cref{tab:bert_imdb}. The results demonstrate the effectiveness of our proposed Gaussian approximations for binary classification. All approximation methods achieve comparable accuracy to the exact binary cross-entropy loss, with differences within the margin of error. The variational and Laplace approximations perform closest to the exact method in terms of NLL, suggesting they provide well-calibrated probability estimates. The moment matching approximation shows slightly higher NLL but still outperforms the naive Gaussian regression on binary labels.

\begin{table}[!ht]
\vspace{-10pt}
\setlength{\tabcolsep}{2.5pt}
\centering
\caption{Predictive performance of a DistilBERT model, fine-tuned on the IMDB dataset using \cref{eqn:approx_map} with binary cross-entropy and various log-likelihood approximations described in \cref{sec:beta_binary}}
\begin{tabular}{ccc}
\hline
Method & ACC & NLL \\
\hline
 exact & 0.913 $\pm$ 0.004 & 0.223 $\pm$ 0.011\\
 gauss & 0.913 $\pm$ 0.002 & 0.523 $\pm$ 0.002\\
 moment & 0.912 $\pm$ 0.003 & 0.282 $\pm$ 0.011\\
 laplace & 0.912 $\pm$ 0.003 & 0.241 $\pm$ 0.008\\
 variational & 0.914 $\pm$ 0.002 & 0.251 $\pm$ 0.009\\
 \hline
\end{tabular}
\label{tab:bert_imdb}
\vspace{-15pt}
\end{table}

\subsection{Gaussian process multiclass classification}
\label{sec:exp_gpc}

\begin{table*}[!ht]
\setlength{\tabcolsep}{3.5pt}
\small
\centering
\vspace{-20pt}
\caption{Performance of various approximations in GP multiclass classification on standard UCI datasets\label{table:gpc}}
\begin{tabular}{cc cccc c cccc c cccc}
\hline
\multirow{2}{*}{Dataset} & \multirow{2}{*}{$N_\text{train}$/$N_\text{test}$/$D$/$K$} & \multicolumn{4}{c}{Error [\%, $\downarrow$]} && \multicolumn{4}{c}{NLL [nat, $\downarrow$]} && \multicolumn{4}{c}{ECE [$\downarrow$]} \\ \cline{3-6} \cline{8-11}  \cline{13-16} 
\multicolumn{1}{c}{}                         &                           & MM-ori & MM      & LM     & VM     && MM-ori & MM     & LM     & VM    && MM-ori & MM     & LM     & VM    \\ \hline
eeg & 10980/4000/14/2 & 16.39 &28.17 &21.54 &23.22 && 0.31 &0.46 &0.50 &0.36 && 0.09 &0.22 &0.19 &0.01  \\
htru2 & 12898/5000/8/2 & 2.02 &2.17 &2.17 &2.02 && 0.07 &0.07 &0.12 &0.07 && 0.04 &0.04 &0.07 &0.04  \\
magic & 14020/5000/10/2 & 12.99 &13.73 &13.96 &13.15 && 0.31 &0.36 &0.42 &0.32 && 0.01 &0.08 &0.15 &0.01  \\
letter & 15000/5000/16/26 & 12.13 &13.21 &16.48 &11.50 && 0.37 &0.39 &1.00 &0.35 && 0.03 &0.03 &0.40 &0.03  \\
drive & 48509/10000/48/11 & 0.78 &0.64 &9.11 &0.46 && 0.02 &0.03 &0.37 &0.02 && 0.05 &0.06 &0.18 &0.05  \\
\hline
\end{tabular}
\end{table*}
\begin{figure*}[!ht]
    \centering
    \vspace{-8pt}
    \includegraphics[width=\linewidth]{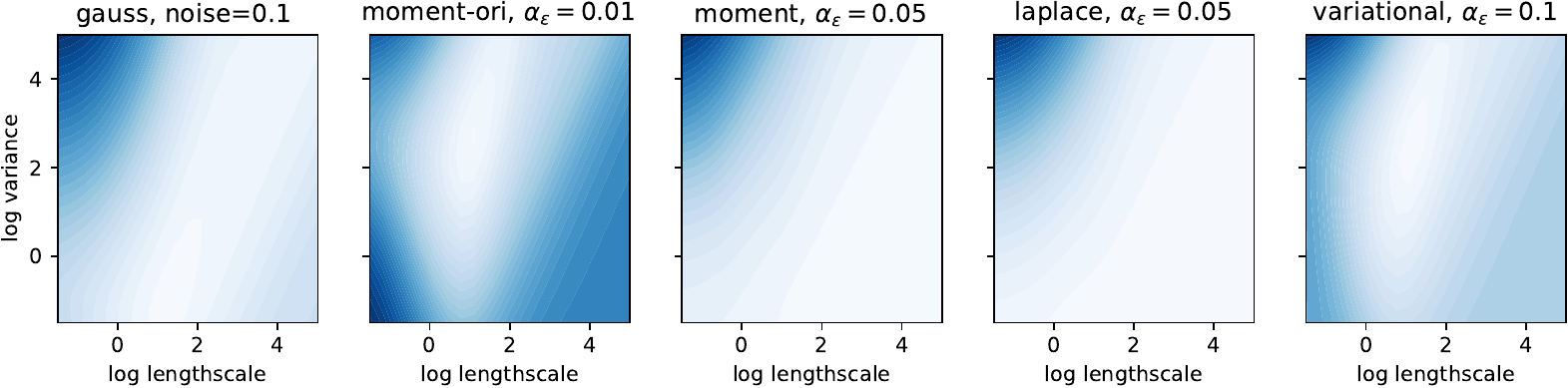}
    \vspace{-20pt}
    \caption{Approximate log marginal likelihood landscapes for different Gaussian approximations of the softmax likelihood on the ionosphere dataset. The x-axis represents the log-lengthscale, and the y-axis represents the log-variance of the exponentiated quadratic kernel. Brighter colours indicate higher likelihood values.}
    \label{fig:gpc_hyper}
    \vspace{-5pt}
\end{figure*}
To evaluate the effectiveness of our approximations in the Bayesian nonparametric settings, we consider Gaussian process multiclass classification tasks on several UCI datasets. 
\Cref{table:gpc} presents the classification error, NLL, and ECE. The errors across ten runs are included in the appendix. Overall, the moment-ori and variational approximations perform best across datasets.
The laplace approximation tends to have higher error rates and poorer NLL and ECE values, particularly on datasets with more classes (e.g., letter and drive). Since \cite{milios2018dirichlet} have presented results comparing moment-ori to label regression and standard posterior approximation schemes like EP and VI, we do not repeat it here. The key benefit of the Gaussian likelihood approximations considered here is that we avoid the need to parameterise and optimise a variational approximation for small to medium-scale datasets, resulting in time savings.

To investigate why seemingly similar Gaussian approximations might yield quantitatively different predictive performance, we plot the approximate log marginal likelihood on the \textit{ionosphere} dataset for various hyperparameters in \cref{fig:gpc_hyper}. We note the similarity in the valley and sharpness of the plots for moment-ori and variational when compared to that of full MCMC or other approximation strategies on the exact likelihood \citep[see e.g.][]{Li24hyper}. Laplace and moment matching schemes result in very different landscapes and tend to give a higher optimal lengthscale and thus smoother predictions.

\vspace{-10pt}
\subsection{Online approximate inference for streaming data}
\label{sec:exp_streaming}
\begin{figure*}[!ht]
    \centering
    \includegraphics[width=\linewidth]{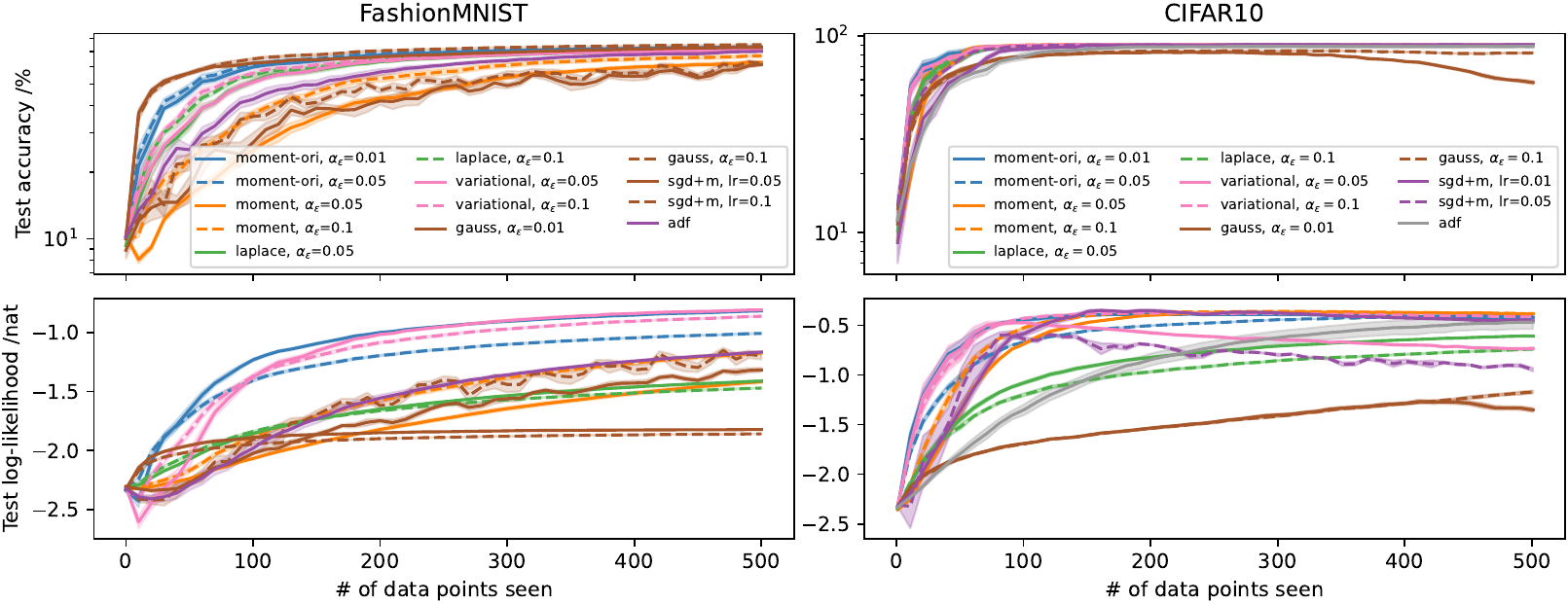}
    \vspace{-20pt}
    \caption{Performance comparison of different approximation methods in online learning scenarios for FashionMNIST (left) and CIFAR-10 (right) datasets. The top row shows test accuracy, and the bottom row shows test log-likelihood, both as functions of the number of data points seen. For FashionMNIST, we use random neural network features, while for CIFAR-10, we use pretrained VGG features.}
    \label{fig:streaming}
    \vspace{-15pt}
\end{figure*}
To evaluate the performance of our approximations in online learning scenarios, we conducted experiments on streaming data for MNIST, FashionMNIST, and CIFAR-10 datasets. This mimics real-world scenarios in which data arrive sequentially and thus efficient posterior updates and predictions are required. For baselines, we compare our methods to Stochastic Gradient Descent with momentum (sgd+m), and Assumed Density Filtering \citep[][ADF]{hernandezlobato}, a Bayesian online learning method that approximates the posterior distribution after each observation. For ADF, we used a correlated Gaussian approximation per output dimension, estimated the log-normaliser of the tilted distribution by simple Monte Carlo, and used damping to avoid invalid updates. For all datasets, we only learn the last layer of the neural networks with a random neural network feature extractor for MNIST/FashionMNIST and pretrained VGG-11 features for CIFAR-10. This setting is relevant in the case where only the last layer of pretrained models is fine-tuned for downstream tasks \citep[e.g][]{titsias2023kalman}. \Cref{fig:streaming} summarises the test performance as a function of the number of data points seen so far.

For FashionMNIST, the moment-ori and variational approximations consistently outperform other approximations in both accuracy and log-likelihood, except gauss in terms of accuracy. The Laplace approximation shows competitive accuracy but lower log-likelihood, indicating potential miscalibration. Both ADF and SGD-m are inferior in terms of performance-sample efficiency. Gaussian regression on the one-hot labels is surprisingly able to achieve high with a small number of samples but does not give calibrated predictions. The hyperparameters of the approximations can affect the performance but do not have a marked impact on the general gaps between the methods in this streaming setting. The results for MNIST are similar and included in the appendix.

For CIFAR-10, the performance gap between different methods is smaller, most likely due to the use of pretrained features. However, the moment-ori and variational approximations still show a slight edge, especially in log-likelihood. Several instances such as variational ($\alpha_\epsilon=0.05)$ and sgd+m exhibit overfitting (or forgetting), indicating the brittleness of SGD in the streaming setting or the approximate posterior corresponding to the likelihood approximations collapsing faster than it should be.

\begin{figure}[!ht]
    \vspace{-8pt}
    \centering
    \includegraphics[width=1.0\linewidth]{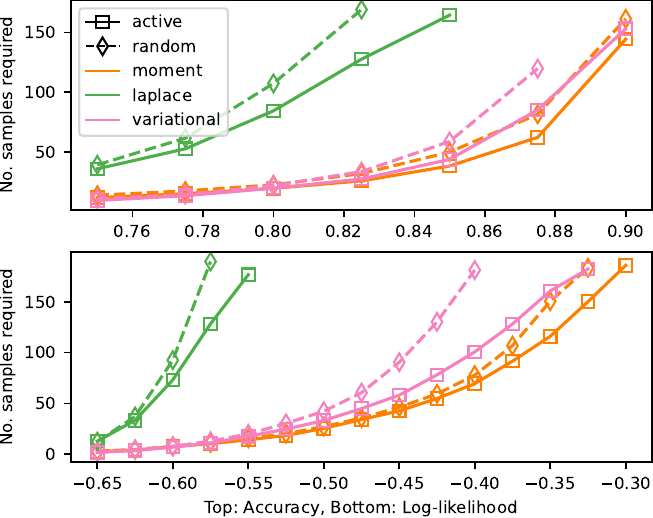}
    \vspace{-15pt}
    \caption{Active learning of a Bayesian logistic regression model on the IMDB dataset.}
    \label{fig:imdb_active_sample}
    \vspace{-10pt}
\end{figure}

Finally, we stress-tested the proposed methods in \cref{sec:beta_binary} for active learning, where Gaussian likelihood approximations enable computationally efficient posterior updates as new data arrive. We considered the IMDB review sentiment dataset, a pool-based setting with a pool set of 5000 instances and a test set with 10000 instances, a Bayesian logistic regression model with features provided by the Gemma-2B model \citep{gemmateam2024gemma2improvingopen}, and an initial training set of 10 instances. At each active learning step, we selected \textit{one} data point from the pool set that has the highest predictive entropy and repeated this for 200 steps and 50 random seeds. The average test accuracy and log-likelihood are collected at each step. The number of samples required to reach various performance levels are reported in \cref{fig:imdb_active_sample}, demonstrating the benefits of active learning and the moment-matching and variational approximations proposed in \cref{sec:beta_binary}. We could not get approximate inference strategies in the exact models, such as ADF and variational Bayes, to work reliably and efficiently in this sequential setting, similar to earlier observations made by \cite{riquelme2018deep}. In contrast, each full active learning run using the proposed Gaussian likelihood approximations takes only seconds on a standard laptop.

\vspace{-5pt}
\section{Related work}
\vspace{-5pt}
Gaussian approximations have been widely used in machine learning to simplify complex distributions, particularly in the context of non-Gaussian likelihoods. Two notable approaches are particularly relevant to our work. \cite{milios2018dirichlet} proposed a Dirichlet-based Gaussian approximation for large-scale calibrated classification, using moment matching in the standard basis to approximate the softmax likelihood. More recently, \cite{hobbhahn2022laplacematching} introduced Laplace matching for fast approximate inference in latent Gaussian models. Our work significantly extends these approaches by (1) introducing and comparing multiple approximation techniques, including moment matching and variational inference in transformed bases; (2) demonstrating applicability to neural network training, broadening the scope beyond Gaussian process models; and (3) showing effectiveness in online and streaming data scenarios;

Turning models with intractable posteriors into approximate ones but with tractable posterior can benefit multiple areas beyond classification \citep{kuss2006gaussian,wilson2016deep,patacchiola2020bayesian}. For example, \cite{titsias2023kalman} employs Kalman filtering on the last layer of neural networks, with a Gaussian likelihood on the one-hot labels, thus needing expensive recalibration. The Gaussian approximations considered here are directly applicable in such settings. In contextual bandits and active learning, Gaussian approximations to the likelihoods enable rapid posterior updates, sidestepping the need for costly approximate inference \citep{riquelme2018deep}. For large pretrained models, Gaussian approximations facilitate efficient fine-tuning of the last linear or GP layer, allowing for quick adaptation to new tasks \citep{Liu23}. In continual learning, these approximations provide a computationally efficient way to maintain and update uncertainty estimates as new data arrives \citep{gal2017deep,BuiNguTur17,titsias2019functional,nguyenLBT18,kapoor21}.

\vspace{-9pt}
\section{Summary}
\vspace{-9pt}

We have introduced a unified and improved approach to approximating non-Gaussian likelihoods with Gaussian distributions that have broad applicability. The proposed methods, based on variational and moment matching strategies, provide a flexible framework for approximating softmax and logistic likelihoods with effectively \textit{no} computational overheads. They achieve comparable or superior performance to exact likelihoods while retaining calibration fidelity in many challenging cases such as GP and neural network classification. The proposed Gaussian approximations also excel in streaming settings where efficient and calibrated posterior updates are crucial.

\subsubsection*{Acknowledgements}
We thank Hippolyt Ritter for the feedback, and acknowledge the support from Meta, Google Cloud, and NCI Australia.

\subsubsection*{References}
\begingroup
\renewcommand{\section}[2]{}%
\bibliography{arxiv}
\endgroup

\appendix
\onecolumn
\aistatstitle{Likelihood approximations via Gaussian approximate inference: \\
Supplementary Materials}
\appendix
\section{Basis transformations and Gaussian approximation}
We provide full derivations of approximations presented in the main text, and additional examples of Laplace, moment and variational matching strategies on other transformed densities.

\subsection{Gamma distribution and $\log$ basis}

If $\omega \sim \text{Gamma}(\alpha, \beta)$ and $\psi = g(\omega) = \log(\omega)$, then $\omega = g^{-1}(\psi) = \exp(\psi)$, $\frac{\partial g^{-1}(\psi)}{\partial \psi} = \exp(\psi)$, and thus $p_{\psi}(\psi) = \text{Gamma}(\exp(\psi); \alpha, \beta) \exp(\psi)$.

\paragraph{Laplace matching}
This was provided by \cite{hobbhahn2022laplacematching}, we repeat their derivation here.
To find the mode, we set the derivative of $\log p_{\psi}(\psi)$ to zero:
$\frac{d}{d\psi} \log p_{\psi}(\psi) = \alpha - \beta \exp(\psi) = 0$. Solving this equation gives the mode: $\psi_{\text{mode}} = \log(\alpha/\beta)$. The second derivative is:
$\frac{d^2}{d\psi^2} \log p_{\psi}(\psi) = -\beta \exp(\psi) = -\alpha$ at the mode. Therefore, the Laplace approximation is:
$q(\psi) = \mathcal{N}(\psi; \log(\alpha/\beta), \alpha^{-1})$

\paragraph{Variational matching}
For $q(\psi) = \Ncal(\psi; \mu, \sigma^2)$, the variational objective is $\mathcal{F}(q) = c - 0.5 \log(\sigma^2) - \alpha \mu + \beta \exp(\mu + 0.5 \sigma^2)$. Setting the gradients wrt.~$\mu$ and $\sigma^2$ to 0 gives $q(\psi) = \Ncal(\psi; \log(\alpha/\beta) - 0.5 \alpha^{-1}, \alpha^{-1})$
\paragraph{Moment matching} We first compute the mean of $\psi$: $\mathbb{E}_\psi[\psi] = \mathbb{E}_\omega[\log(\omega)] = \Psi(\alpha) - \log(\beta)$ where $\Psi$ is the digamma function. Similarly, the variance $\mathbb{V}_\psi[\psi] = \Psi^{1}(\alpha)$ where $\Psi^1$ is the trigamma function. Therefore, $q(\psi) = \Ncal(\psi; \Psi(\alpha) - \log(\beta), \Psi^1(\alpha))$.

We visually compare the above approximations and the log-normal approximation in \citep{milios2018dirichlet} in \cref{fig:beta_approx}.

\begin{figure*}[!th]
    \vspace{-8pt}
    \includegraphics[width=\linewidth]{{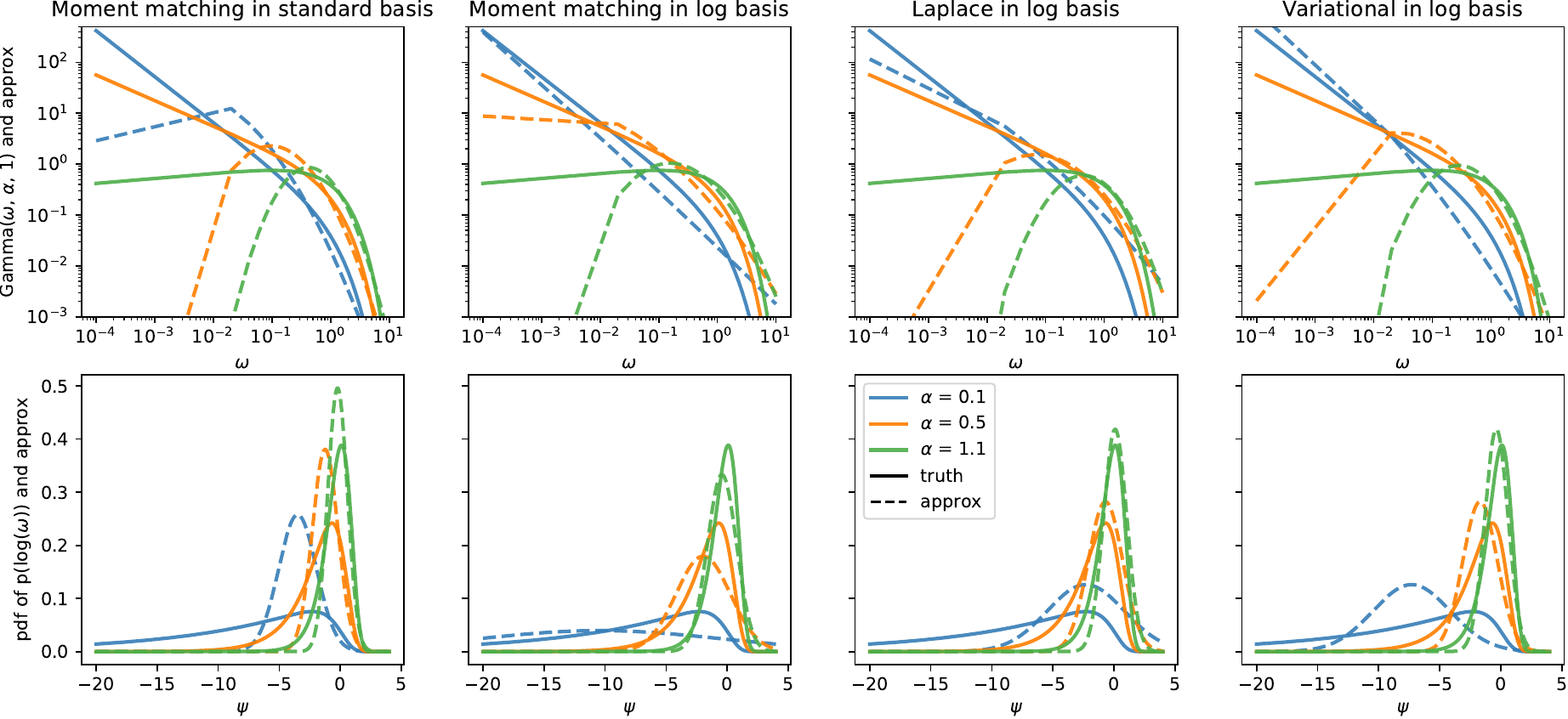}}
    \caption{Approximations of Gamma distributions using log-Normal moment matching \citep[][first column]{milios2018dirichlet}, Laplace approximation in log basis \citep[][third column]{hobbhahn2022laplacematching}, and Gaussian moment matching (second column) and variational matching (fourth column). The first row shows the Gamma pdfs and their log-normal approximations. The second row shows the log-transformed densities and their approximations.}
    \label{fig:gamma_approx_viz}
\end{figure*}

\subsection{Beta distribution, $\mathrm{logit}$ basis}
If $\omega \sim \mathrm{Beta}(\alpha, \beta)$ and $\psi = g(\omega) = \mathrm{logit}(\omega) = \log(\frac{\omega}{1-\omega})$, then $\omega = g^{-1}(\psi) = \phi(\psi)$, $\frac{\partial g^{-1}(\psi)}{\partial \psi} = \phi(\psi)(1-\phi(\psi))$, and $p_{\psi}(\psi) = \mathrm{Beta}(\phi(\psi); \alpha, \beta) \phi(\psi)(1-\phi(\psi))$, with $\phi$ is the logistic sigmoid function.

\paragraph{Laplace matching}
As derived in \citep{hobbhahn2022laplacematching}, the Laplace matching approximation is:
$q(\psi) = \mathcal{N}(\psi; \log(\alpha / \beta), (\alpha + \beta) / (\alpha \beta)$.

\paragraph{Variational matching}
For $q(\psi) = \mathcal{N}(\psi; \mu, \sigma^2)$, the variational objective is:
$\mathcal{F}(q) = c - 0.5 \log(\sigma^2) + \beta \mu + (\alpha + \beta) \mathbb{E}_q[\log(1 + \exp(-\psi))]$.
This does not have a closed-form solution but we can use numerical optimisation and the reparameterisation trick \citep{rezende14,kingma2013auto} to estimate the variational parameters $\mu$ and $\sigma^2$.

\paragraph{Moment matching}
The derivation of the mean and variance of $\psi$ is more involved but gives close-form solutions, yielding  $q(\psi) = \Ncal(\psi; \Psi(\alpha) - \Psi(\beta), \Psi^1(\alpha) + \Psi^1(\beta))$. 

\begin{figure}[!ht]
    \centering
    \includegraphics[width=\linewidth]{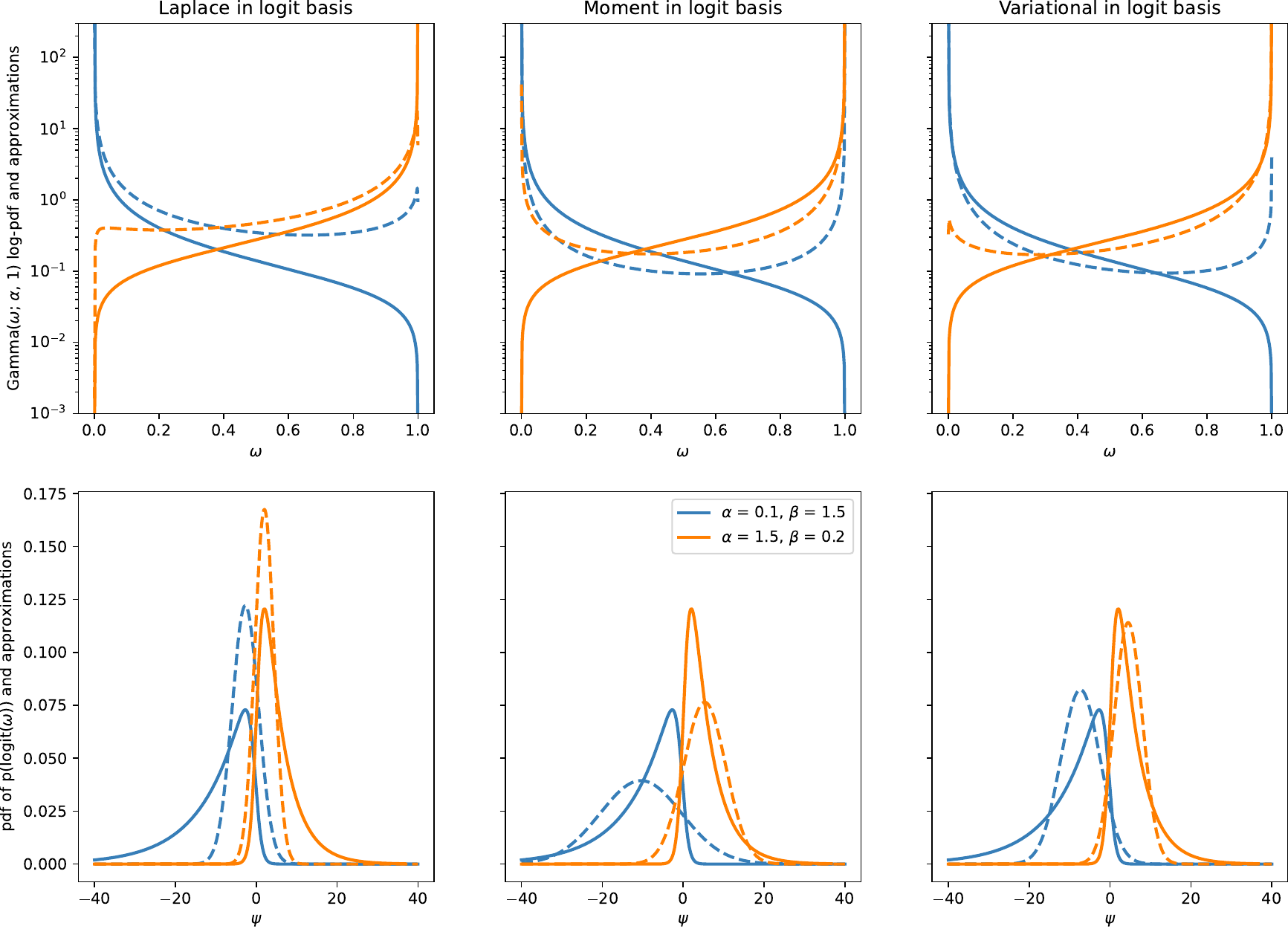}
    \caption{Approximations of Beta distributions Laplace approximation \citep[][first column]{hobbhahn2022laplacematching}, and Gaussian moment matching (second column) and variational matching (third column). The first row shows the Beta pdfs and their logit$^{-1}$-transformed Normal approximations. The second row shows the logit-transformed densities and their approximations.}
    \label{fig:beta_approx}
\end{figure}

We visually compare the above approximations in \cref{fig:beta_approx}.

\subsection{Exponential distribution and $\log$ basis}
If $\omega \sim \mathrm{Exp}(\lambda)$ and $\psi = g(\omega) = \log(\omega)$, then $p_{\psi}(\psi) = \mathrm{Exp}(\exp(\psi); \lambda) \exp(\psi) = \lambda \exp(\psi - \lambda \exp(\psi))$. While this can be considered a special log-transformed Gamma distribution case, we still provide the direct derivations for completeness.

\paragraph{Laplace matching} We can find the mode and Hessian of $\log p_{\psi}(\psi) = \psi - \lambda \exp(\psi)$ analytically to obtain $q(\psi) = \mathcal{N}(\psi; -\log(\lambda), 1)$ \citep{hobbhahn2022laplacematching}.

\paragraph{Variational matching}
For $q(\psi) = \mathcal{N}(\psi; \mu, \sigma^2)$, we minimize $\mathcal{F}(q) = c - 0.5 \log(\sigma^2) - \mu + \lambda\exp(\mu + 0.5\sigma^2)$. Setting the partial derivatives wrt $\mu$ and $\sigma^2$ to zero gives the variational matching approximation
$q(\psi) = \mathcal{N}(\psi; -\log(\lambda) - 0.5, 1)$.

\paragraph{Moment matching}
Interestingly, the first two moments of $\psi$ are available in closed-form, leading to $q(\psi) = \mathcal{N}(\psi; -\log(\lambda) - \mathrm{em}, \pi^2/6)$, where $\mathrm{em} = 0.57721...$ is the Euler-Mascheroni constant. We note that the variance of the moment-matched Gaussian is $\pi^2/6 \approx 1.645$ and larger than that of Laplace or variational matching.

\subsection{Inverse Gamma distribution and $\log$ basis}

If $\omega \sim \text{InvGamma}(\alpha, \beta)$ and $\psi = g(\omega) = \log(\omega)$, then $\omega = g^{-1}(\psi) = \exp(\psi)$, $\frac{\partial g^{-1}(\psi)}{\partial \psi} = \exp(\psi)$, and thus $p_{\psi}(\psi) = \text{InvGamma}(\exp(\psi); \alpha, \beta) \exp(\psi)$.

\paragraph{Laplace matching}
The log-density is $\log p_{\psi}(\psi) = c - \alpha\psi - \beta \exp(-\psi)$. The mode is found by setting the derivative to zero:
$-\alpha + \beta \exp(-\psi) = 0$.
Solving this gives the mode at $\psi_{\text{mode}} = \log(\beta / \alpha)$. The second derivative at this point is $-\alpha$. Therefore, the Laplace approximation is:
$q(\psi) = \mathcal{N}(\psi; \log(\beta / \alpha), 1/\alpha)$.

\paragraph{Variational matching}
For $q(\psi) = \mathcal{N}(\psi; \mu, \sigma^2)$, the variational objective is:
$\mathcal{F}(q) = c - 0.5 \log(\sigma^2) - \alpha\mu + \beta \mathbb{E}_q[\exp(-\psi)] = c - 0.5 \log(\sigma^2) - \alpha\mu - \beta \exp(-\mu + 0.5\sigma^2)$.
Setting the gradients to zero:
$\frac{\partial \mathcal{F}}{\partial \mu} = -\alpha + \beta \exp(-\mu + 0.5\sigma^2) = 0$, and 
$\frac{\partial \mathcal{F}}{\partial \sigma^2} = -\frac{1}{2\sigma^2} - \frac{\beta}{2} \exp(-\mu + 0.5\sigma^2) = 0$.
Solving these equations gives:
$\mu = \log(\beta / \alpha) + 0.5/\alpha$
$\sigma^2 = 1/\alpha$.
Therefore, the variational matching approximation is:
$q(\psi) = \mathcal{N}(\psi; \log(\beta / \alpha) + 0.5/\alpha, 1/\alpha)$.

\paragraph{Moment matching}
For the inverse Gamma distribution, $\mathbb{E}[\log(\omega)] = \log(\beta) - \Psi(\alpha)$, where $\Psi$ is the digamma function. The variance of $\log(\omega)$ is $\Psi^1(\alpha)$, where $\Psi^1$ is the trigamma function. Therefore, the moment matching approximation is:
$q(\psi) = \mathcal{N}(\psi; \log(\beta) - \Psi(\alpha), \Psi^1(\alpha))$.

\subsection{Chi-squared distribution and $\log$ basis}
If $\omega \sim \chi^2(k)$ (chi-squared with $k$ degrees of freedom) and $\psi = g(\omega) = \log(\omega)$, then $\omega = g^{-1}(\psi) = \exp(\psi)$, $\frac{\partial g^{-1}(\psi)}{\partial \psi} = \exp(\psi)$, and thus $p_{\psi}(\psi) = \chi^2(\exp(\psi); k) \exp(\psi)$. The chi-squared distribution is a special case of Gamma distribution, that is $\omega \sim \chi^2(k)$ is equivalent to $\omega \sim \text{Gamma}(k/2, 1/2)$.

\paragraph{Laplace matching}
The log-density is $\log p_{\psi}(\psi) = c + 0.5k\psi - 0.5\exp(\psi)$. The mode is found by setting the derivative to zero:
$0.5 (k - \exp(\psi)) = 0$. Solving this gives the mode at $\psi_{\text{mode}} = \log(k)$. The second derivative at this point is $-k/2$. Therefore, the Laplace approximation is:
$q(\psi) = \mathcal{N}(\psi; \log(k), 2/k)$.

\paragraph{Variational matching}
For $q(\psi) = \mathcal{N}(\psi; \mu, \sigma^2)$, the variational objective is:
$\mathcal{F}(q) = c - 0.5 \log(\sigma^2) + 0.5k\mu - 0.5\mathbb{E}_q[\exp(\psi)] = c - 0.5 \log(\sigma^2) + 0.5k\mu - 0.5\exp(\mu + 0.5\sigma^2)$.
Setting the gradients to zero:
$\frac{\partial \mathcal{F}}{\partial \mu} = k/2 - 0.5\exp(\mu + 0.5\sigma^2) = 0$ and 
$\frac{\partial \mathcal{F}}{\partial \sigma^2} = -\frac{1}{2\sigma^2} - \frac{1}{4} \exp(\mu + 0.5\sigma^2) = 0$.
Solving these equations gives:
$\mu = \log(k) - 1/k$
$\sigma^2 = 2/k$
Therefore, the variational matching approximation is:
$q(\psi) = \mathcal{N}(\psi; \log(k) - 1/k, 2/k)$.

\paragraph{Moment matching}
For the chi-squared distribution, $\mathbb{E}[\log(\omega)] = \Psi(k/2) + \log(2)$, where $\Psi$ is the digamma function. The variance of $\log(\omega)$ is $\Psi^1(k/2)$, where $\Psi^1$ is the trigamma function. Therefore, the moment matching approximation is:
$q(\psi) = \mathcal{N}(\psi; \Psi(k/2) + \log(2), \Psi^1(k/2))$.

\newpage
\section{Parameterising variational distributions with likelihood approximations}
Instead of performing exact inference in the approximate model, we can use the approximate Gaussian likelihood to construct the approximate posterior and then perform approximate inference under the exact model. The benefits are two-fold. First, we have a principled way to initialise the approximate posterior. Second, the hyperparameters that are used to construct the auxiliary observations and heteroscedastic noise can be treated as approximation parameters and optimised, together with the model parameters and other variational parameters, using the variational objective. The only variational parameters in this scheme are $\alpha_\epsilon$, or $\alpha_\epsilon$ and $\beta_\epsilon$ used in the likelihood approximations.

\begin{figure}[!ht]
    \centering
    \includegraphics[width=\linewidth]{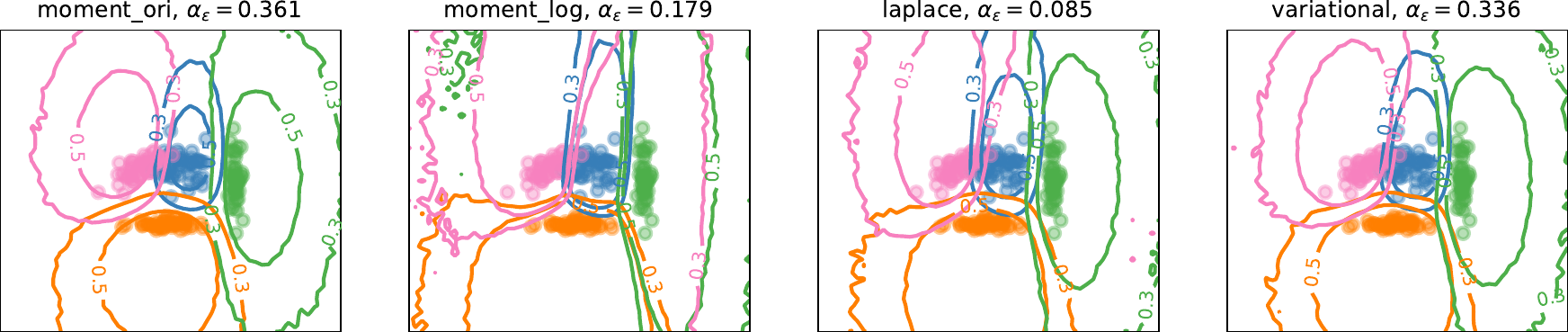}
    \caption{Variational GP classification on a toy dataset with four classes. Each subtitle has the name of the Gaussian likelihood approximation used to construct the variational approximation, and $\alpha_\epsilon$ found by optimising the variational lower bound.}
    \label{fig:fourclass_vdgpc}
\end{figure}

We will use Gaussian process multiclass classification to illustrate this. Consider a classification problem with $K$ classes and $N$ training data points $\{\xvec_n, y_n\}_{n=1}^{N}$, and assume there are $K$ latent functions $\{f_k\}_{k=1}^{K}$ that maps inputs to class logits. We place independent GP priors over the latent functions and combine them with the likelihood to find the inverse probability,
\begin{align*}
    p(f_{1:K} | \Dcal, \theta) = \frac{p(f_{1:K} | \theta) p(\Dcal | f_{1:K})} {p(\Dcal | \theta)} \quad \text{where} \quad p(\Dcal | \theta) = \int_f p(f_{1:K} | \theta) p(\Dcal | f_{1:K}), \; \text{and} \; p(f_{k}) = \mathcal{GP}(f_k; m(\cdot), k(\cdot,\cdot)),
\end{align*}
where $\theta$ is the kernel hyperparameters. We can approximate the posterior above by a variational distribution,
\begin{align*}
    q(f_{1:K} | \theta, \alpha_\epsilon) \propto p(f_{1:K} | \theta) \prod_k \prod_n \Ncal(\tilde{y}_{k,n}; f_{k,n}, \tilde{v}_{k,n}) = p(f_{1:K, \neq \mathbf{f}} | \theta, \mathbf{f}_{1:K}) \prod_k q(\mathbf{f}_k | \theta, \alpha_\epsilon),
\end{align*}
where $\tilde{y}_{k,n}$ and $\tilde{v}_{k,n}$ are the mean and variance of the Gaussian approximations as discussed in \cref{sec:gamma_multiclass}. The resulting lower bound to the log marginal likelihood is
\begin{align*}
    \mathcal{L}(\theta, \alpha_\epsilon) = -\sum_k \mathrm{KL}[q(\mathbf{f}_k | \theta, \alpha_\epsilon) \; || \; p(\mathbf{f}_k | \theta)] + \sum_n \int_{\mathbf{f}_{1:K}} q(\mathbf{f}_{1:K} | \theta, \alpha_\epsilon) \log p(y_n | f_{1:K}).
\end{align*}
This is identical to the standard variational Gaussian process classification, except that the variational posterior is now provided as the exact posterior in a heteroscedastic Gaussian process regression with observations $\mathbf{\tilde{y}}$ and noise $\mathbf{\tilde{v}}$. We will use the above objective to optimise for the kernel hyperparameters $\theta$ and the variational parameter $\alpha_\epsilon$. \Cref{fig:fourclass_vdgpc} demonstrates the predictions made by this variational scheme on a toy dataset.

In addition, we can use this scheme to initialise the variational approximation in standard variational GP classification and speed up convergence. We plan to explore this direction in future work. 




\newpage
\section{Full experimental results}

\subsection{Training neural networks using regularised maximum likelihood}

\subsubsection{Convolutional neural networks and MNIST}
We train a convolutional neural network on MNIST using the exact softmax likelihood or the proposed likelihood approximations. The network has two convolutional layers that have 32 and 64 3x3 filters with a stride of 1, respectively. They are followed by a max-pooling operation and two fully connected layers. We use dropout after max-pooling and the first fully connected layers, and rectified linear nonlinearity. We train for 10 epochs, use an adaptive learning rate, and decay the learning rate after each epoch. 
\begin{figure}[!ht]
    \vspace{-10pt}
    \centering
    \includegraphics[width=0.78\linewidth]{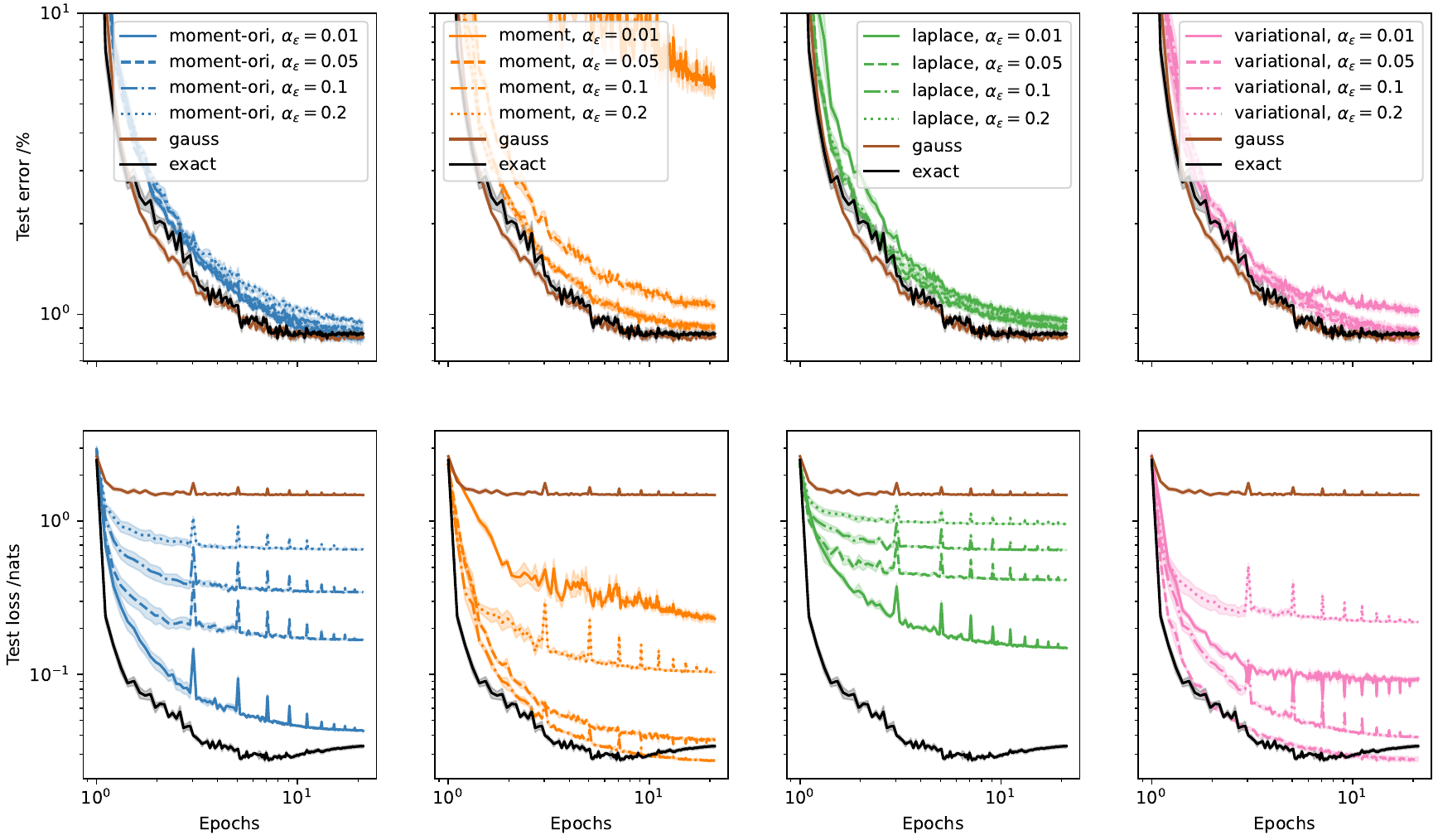}
    \vspace{-10pt}
    \caption{Performance of a CNN on the MNIST test set, trained using the cross-entropy loss and various Gaussian approximations. We vary the hyperparameter $\alpha_\epsilon$ of the Gamma approximation discussed in \cref{sec:gamma_multiclass}.}
    \label{fig:mnist_all}
    \vspace{-10pt}
\end{figure}

\begin{figure}[!ht]
    \vspace{-10pt}
    \centering
    \includegraphics[width=0.5\linewidth]{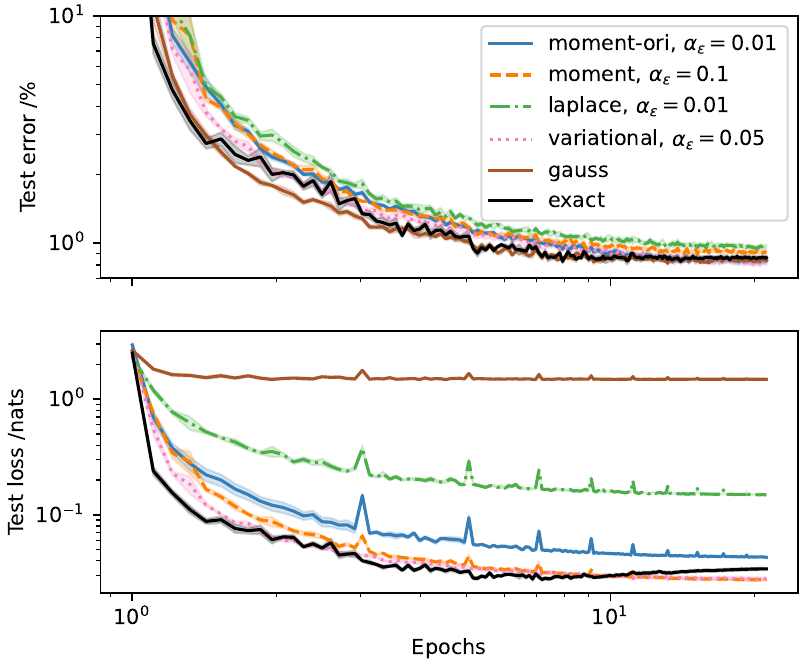}
    \caption{Performance of a CNN on the MNIST test set, trained using the cross-entropy loss and various Gaussian approximations.
    \label{fig:mnist_best}}
    \vspace{-10pt}
    \vspace{-10pt}
\end{figure}

The test performances during training for various methods and hyperparameters are shown in \cref{fig:mnist_all}. The errors are similar across different hyperparameters whilst the test likelihoods are sensitive to $\alpha_\epsilon$. Similar to \cite{milios2018dirichlet}, we note that the test likelihood is correlated with the train likelihood, so we can use the train likelihood to select $\alpha_\epsilon$ for the approximation. We pick the best version for each method and plot them in \cref{fig:mnist_best}. We note that $\alpha_\epsilon$ between 0.1 and 0.2 seems to work well across all datasets and architectures we have tried.


\newpage
\subsubsection{ResNet-18 and VGG-11 on CIFAR-10}
Next, we train ResNet-18 and VGG-11 networks on the CIFAR-10 dataset using the exact cross-entropy loss and the approximations discussed in \cref{sec:gamma_multiclass}. The networks were trained for 200 epochs with SGD with momentum and the learning rate drops every 40 epochs. We evaluate the in-distribution performance on the CIFAR-10 test set, and out-of-distribution detection (using the maximum logit value) on the CIFAR-100 test set (near OOD) and the SVHN test test (far OOD). Each experiment was repeated 10 times. We report the mean and standard errors of all evaluation metrics across various hyperparameters in \cref{tab:cifar10_full}. We note that $\alpha_\epsilon$ between 0.1 and 0.2 seems to work well across all datasets and architectures we have tried.
\begin{table*}[!ht]
\setlength{\tabcolsep}{3.5pt}
\small
\centering
\caption{Predictive performance of VGG-11 and ResNet-18 on CIFAR-10, trained with the multiclass cross-entropy loss and various Gaussian approximations. The results are means and standard errors over 10 runs.\label{tab:cifar10_full}}
\begin{tabular}{ccccccc}
\hline
Net & Method (and $\alpha_\epsilon)$ & Accuracy [\%, $\uparrow$] & NLL [nat, $\downarrow$] & ECE [$\downarrow$] &  CIFAR-100 AUC [$\uparrow$] &  SVHN AUC [$\uparrow$] \\ \hline
\multirow{14}{*}{vgg11} & exact & 91.045 $\pm$ 0.049 & 0.360 $\pm$ 0.003 & 0.057 $\pm$ 0.001 & 0.848 $\pm$ 0.000 & 0.858 $\pm$ 0.003\\
& gauss & 88.625 $\pm$ 0.066 & 1.598 $\pm$ 0.000 & 0.670 $\pm$ 0.001 & 0.806 $\pm$ 0.001 & 0.835 $\pm$ 0.006\\
& moment-ori 0.01 & 91.333 $\pm$ 0.037 & 0.362 $\pm$ 0.002 & 0.046 $\pm$ 0.000 & 0.801 $\pm$ 0.001 & 0.826 $\pm$ 0.005\\
& moment-ori 0.05 & 91.098 $\pm$ 0.066 & 0.395 $\pm$ 0.002 & 0.081 $\pm$ 0.001 & 0.807 $\pm$ 0.001 & 0.846 $\pm$ 0.004\\
& moment-ori 0.1 & 91.004 $\pm$ 0.057 & 0.537 $\pm$ 0.001 & 0.213 $\pm$ 0.001 & 0.808 $\pm$ 0.001 & 0.845 $\pm$ 0.004\\
& moment 0.01 & 40.290 $\pm$ 0.604 & 1.663 $\pm$ 0.009 & 0.163 $\pm$ 0.004 & 0.655 $\pm$ 0.003 & 0.587 $\pm$ 0.015\\
& moment 0.05 & 86.498 $\pm$ 0.085 & 0.616 $\pm$ 0.003 & 0.057 $\pm$ 0.001 & 0.755 $\pm$ 0.001 & 0.845 $\pm$ 0.006\\
& moment 0.1 & 89.664 $\pm$ 0.059 & 0.457 $\pm$ 0.003 & 0.067 $\pm$ 0.001 & 0.803 $\pm$ 0.001 & 0.847 $\pm$ 0.003\\
& laplace 0.01 & 87.803 $\pm$ 0.050 & 0.417 $\pm$ 0.001 & 0.080 $\pm$ 0.001 & 0.806 $\pm$ 0.001 & 0.842 $\pm$ 0.004\\
& laplace 0.05 & 89.630 $\pm$ 0.049 & 0.604 $\pm$ 0.001 & 0.254 $\pm$ 0.001 & 0.818 $\pm$ 0.001 & 0.843 $\pm$ 0.004\\
& laplace 0.1 & 89.603 $\pm$ 0.043 & 0.817 $\pm$ 0.001 & 0.390 $\pm$ 0.001 & 0.813 $\pm$ 0.001 & 0.857 $\pm$ 0.005\\
& variational 0.01 & 88.476 $\pm$ 0.064 & 1.117 $\pm$ 0.005 & 0.241 $\pm$ 0.002 & 0.683 $\pm$ 0.001 & 0.796 $\pm$ 0.007\\
& variational 0.05 & 91.082 $\pm$ 0.059 & 0.628 $\pm$ 0.003 & 0.076 $\pm$ 0.001 & 0.800 $\pm$ 0.001 & 0.835 $\pm$ 0.005\\
& variational 0.1 & 91.216 $\pm$ 0.049 & 0.366 $\pm$ 0.002 & 0.051 $\pm$ 0.000 & 0.808 $\pm$ 0.001 & 0.837 $\pm$ 0.005\\
\hline
\multirow{14}{*}{resnet18} & exact & 93.525 $\pm$ 0.034 & 0.247 $\pm$ 0.001 & 0.040 $\pm$ 0.000 & 0.875 $\pm$ 0.001 & 0.881 $\pm$ 0.003\\
& gauss & 91.155 $\pm$ 0.042 & 1.562 $\pm$ 0.001 & 0.690 $\pm$ 0.000 & 0.795 $\pm$ 0.002 & 0.788 $\pm$ 0.012\\
& moment-ori 0.01 & 93.789 $\pm$ 0.053 & 0.284 $\pm$ 0.002 & 0.034 $\pm$ 0.000 & 0.788 $\pm$ 0.002 & 0.782 $\pm$ 0.010\\
& moment-ori 0.05 & 93.789 $\pm$ 0.058 & 0.320 $\pm$ 0.001 & 0.087 $\pm$ 0.001 & 0.784 $\pm$ 0.002 & 0.767 $\pm$ 0.011\\
& moment-ori 0.1 & 93.642 $\pm$ 0.044 & 0.465 $\pm$ 0.001 & 0.217 $\pm$ 0.001 & 0.784 $\pm$ 0.001 & 0.782 $\pm$ 0.008\\
& moment 0.01 & 37.469 $\pm$ 0.295 & 1.780 $\pm$ 0.005 & 0.166 $\pm$ 0.002 & 0.631 $\pm$ 0.002 & 0.454 $\pm$ 0.009\\
& moment 0.05 & 88.479 $\pm$ 0.076 & 0.516 $\pm$ 0.002 & 0.049 $\pm$ 0.001 & 0.772 $\pm$ 0.001 & 0.850 $\pm$ 0.009\\
& moment 0.1 & 92.350 $\pm$ 0.053 & 0.365 $\pm$ 0.003 & 0.050 $\pm$ 0.001 & 0.809 $\pm$ 0.001 & 0.850 $\pm$ 0.011\\
& laplace 0.01 & 90.546 $\pm$ 0.077 & 0.339 $\pm$ 0.002 & 0.072 $\pm$ 0.001 & 0.817 $\pm$ 0.001 & 0.847 $\pm$ 0.008\\
& laplace 0.05 & 92.239 $\pm$ 0.037 & 0.535 $\pm$ 0.001 & 0.251 $\pm$ 0.000 & 0.798 $\pm$ 0.001 & 0.790 $\pm$ 0.012\\
& laplace 0.1 & 92.238 $\pm$ 0.050 & 0.751 $\pm$ 0.001 & 0.394 $\pm$ 0.000 & 0.792 $\pm$ 0.002 & 0.779 $\pm$ 0.009\\
& variational 0.01 & 91.167 $\pm$ 0.059 & 0.868 $\pm$ 0.005 & 0.271 $\pm$ 0.003 & 0.695 $\pm$ 0.002 & 0.777 $\pm$ 0.012\\
& variational 0.05 & 93.525 $\pm$ 0.024 & 0.488 $\pm$ 0.002 & 0.057 $\pm$ 0.000 & 0.806 $\pm$ 0.001 & 0.810 $\pm$ 0.014\\
& variational 0.1 & 93.698 $\pm$ 0.048 & 0.291 $\pm$ 0.002 & 0.040 $\pm$ 0.000 & 0.796 $\pm$ 0.001 & 0.811 $\pm$ 0.012\\
\hline
\end{tabular}
\end{table*}

\newpage
\subsubsection{DistilBERT and IMDB review sentiment dataset}
\begin{figure}[!ht]
    \centering
    \includegraphics[width=0.6\linewidth]{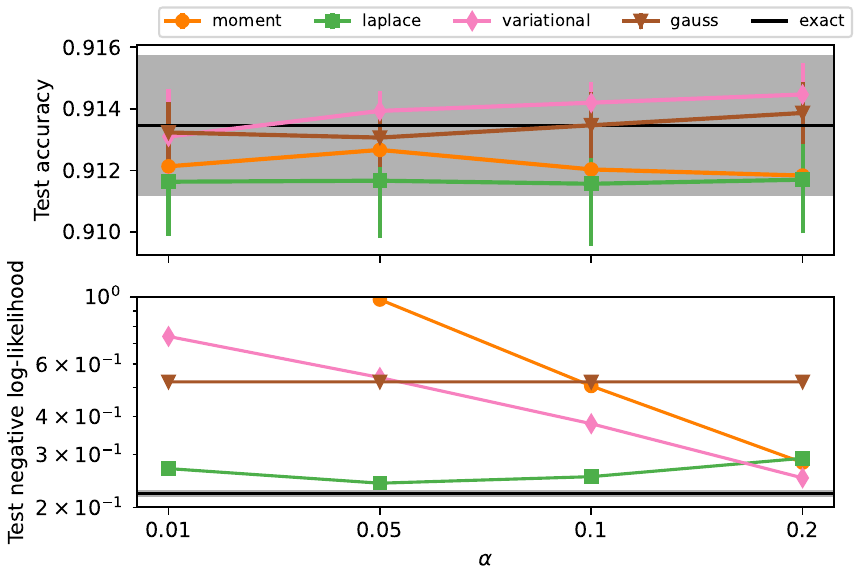}
    \caption{Performance of DistilBERT on the IMDB sentiment dataset using the exact binary cross-entropy loss, least squares on raw labels or the Gaussian approximations discussed in \cref{sec:beta_binary}.}
    \label{fig:distilbert_hypers}
\end{figure}

Next, we finetune the pre-trained DistilBERT model \citep{sanh2019distilbert} on the IMDB review sentiment dataset using the exact binary cross-entropy loss or the least-squares loss corresponding to the approximations in \cref{sec:beta_binary}. We split the dataset into 35000 instances for training, 5000 instances for validation and 100000 instances for evaluation. All layers of the model were fine-tuned for 2 epochs using Adam with weight decay. We repeat the experiments 5 times. The test performances for various hyperparameters are shown in \cref{fig:distilbert_hypers}. Here we set $\alpha_\epsilon = \beta_\epsilon$. Similar to earlier observations, the test accuracies are similar across hyperparameters, whilst the test log-likelihood is sensitive to this value. We note that $\alpha_\epsilon$ between 0.1 and 0.2 seems to work well across all datasets and architectures we have tried.

\newpage

\begin{landscape}
\subsection{Gaussian process classification}
We report the test performance on 20 train/test splits of five UCI datasets trained with collapsed sparse variational process regression \citep{titsias09a} and 200 inducing points in \cref{fig:gpc_full}.
We note that $\alpha_\epsilon$ between 0.1 and 0.2 seems to work well across all datasets and architectures we have tried. Similar to \cite{milios2018dirichlet}, we select $\alpha_\epsilon$ based on the train set, and report the best numbers for each method and dataset in \cref{fig:gpc_best} and \cref{tab:gpc_best_full}.
\begin{figure}[!ht]
    \centering
    \includegraphics[width=\linewidth]{{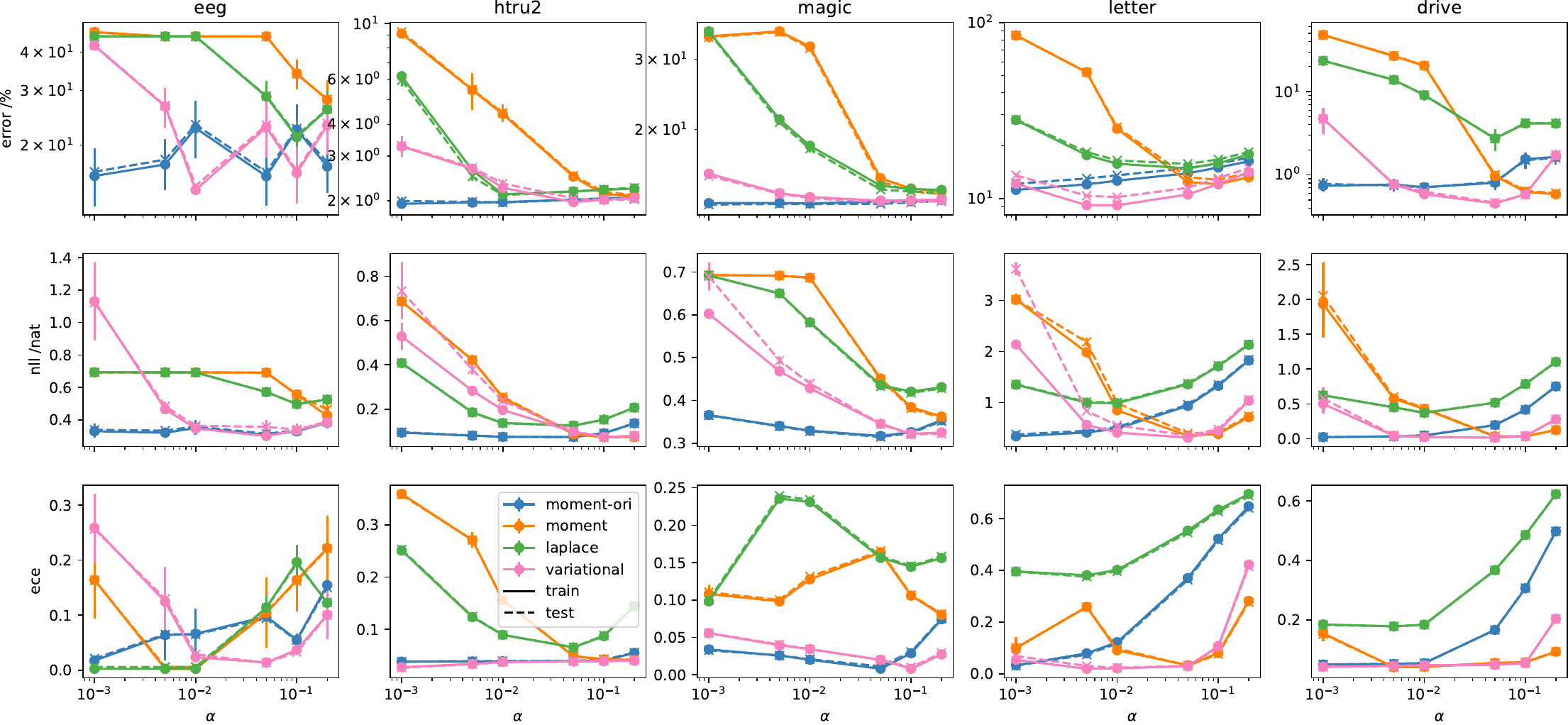}}
    \caption{Gaussian process classification performance using the approximations discussed in \cref{sec:gamma_multiclass}. These approximations allow GP classification to be treated as GP regression with heteroscedastic noise.}
    \label{fig:gpc_full}
\end{figure}
\end{landscape}

\begin{landscape}
\begin{table*}[!ht]
\scriptsize
\caption{GP multiclass classification performance on various UCI datasets using the approximations discussed in \cref{sec:gamma_multiclass}. For figure fanatics, \cref{fig:gpc_best} provides a visual comparison of these numbers.\label{tab:gpc_best_full}}
\begin{tabular}{c cccc | cccc | cccc}
\hline
\multicolumn{1}{c}{\multirow{2}{*}{Dataset}}  & \multicolumn{4}{c}{Error [\%, $\downarrow$]} & \multicolumn{4}{c}{NLL [nat, $\downarrow$]} & \multicolumn{4}{c}{ECE [$\downarrow$]} \\ \cline{2-5} \cline{6-9}  \cline{10-13} 
\multicolumn{1}{c}{}                                                 & MM-ori  & MM      & LA     & VI     & MM-ori & MM     & LA     & VI    & MM-ori    & MM     & LA     & VI    \\ \hline
eeg & 16.39 $\pm$ 3.02 &28.17 $\pm$ 4.31 &21.54 $\pm$ 2.58 &23.22 $\pm$ 4.52 & 0.31 $\pm$ 0.01 &0.46 $\pm$ 0.03 &0.50 $\pm$ 0.01 &0.36 $\pm$ 0.04 & 0.09 $\pm$ 0.04 &0.22 $\pm$ 0.06 &0.19 $\pm$ 0.03 &0.01 $\pm$ 0.00  \\
htru2 & 2.02 $\pm$ 0.06 &2.17 $\pm$ 0.07 &2.17 $\pm$ 0.06 &2.02 $\pm$ 0.06 & 0.07 $\pm$ 0.00 &0.07 $\pm$ 0.00 &0.12 $\pm$ 0.00 &0.07 $\pm$ 0.00 & 0.04 $\pm$ 0.00 &0.04 $\pm$ 0.00 &0.07 $\pm$ 0.00 &0.04 $\pm$ 0.00  \\
magic & 12.99 $\pm$ 0.11 &13.73 $\pm$ 0.16 &13.96 $\pm$ 0.14 &13.15 $\pm$ 0.13 & 0.31 $\pm$ 0.00 &0.36 $\pm$ 0.00 &0.42 $\pm$ 0.00 &0.32 $\pm$ 0.00 & 0.01 $\pm$ 0.00 &0.08 $\pm$ 0.00 &0.15 $\pm$ 0.00 &0.01 $\pm$ 0.00  \\
letter & 12.13 $\pm$ 0.17 &13.21 $\pm$ 0.10 &16.48 $\pm$ 0.13 &11.50 $\pm$ 0.13 & 0.37 $\pm$ 0.01 &0.39 $\pm$ 0.00 &1.00 $\pm$ 0.00 &0.35 $\pm$ 0.00 & 0.03 $\pm$ 0.00 &0.03 $\pm$ 0.00 &0.40 $\pm$ 0.00 &0.03 $\pm$ 0.00  \\
drive & 0.78 $\pm$ 0.14 &0.64 $\pm$ 0.03 &9.11 $\pm$ 1.36 &0.46 $\pm$ 0.04 & 0.02 $\pm$ 0.00 &0.03 $\pm$ 0.00 &0.37 $\pm$ 0.02 &0.02 $\pm$ 0.00 & 0.05 $\pm$ 0.00 &0.06 $\pm$ 0.00 &0.18 $\pm$ 0.00 &0.05 $\pm$ 0.00  \\
\hline
\end{tabular}
\end{table*}

\begin{figure}[!ht]
    \centering
    \includegraphics[width=\linewidth]{{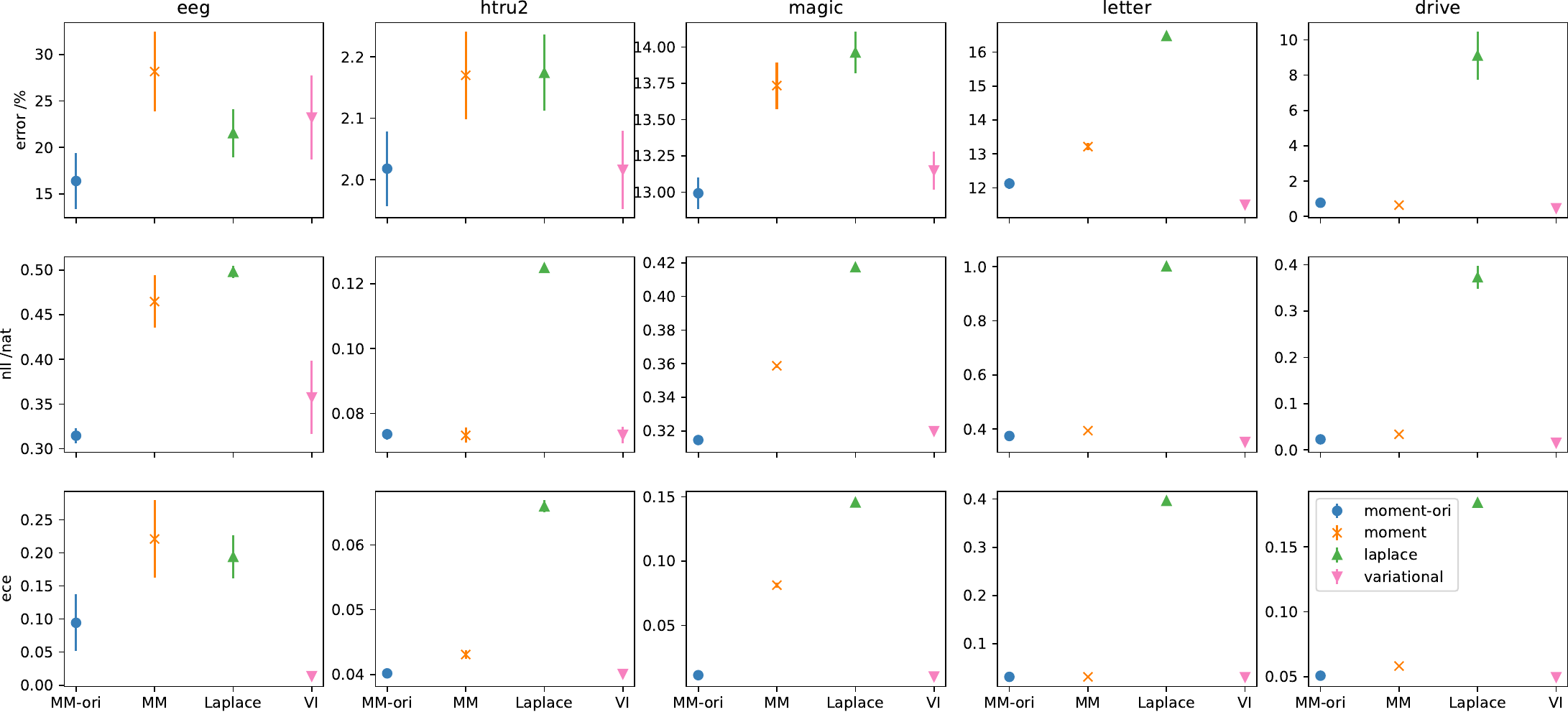}}
    \caption{GP multiclass classification performance on various UCI datasets using the approximations discussed in \cref{sec:gamma_multiclass}. For table fanatics, the raw numbers are shown in \cref{tab:gpc_best_full}.}
    \label{fig:gpc_best}
\end{figure}

\end{landscape}

\newpage
\subsection{Online inference with streaming data}
\begin{figure}[!ht]
    \centering
    \includegraphics[width=\linewidth]{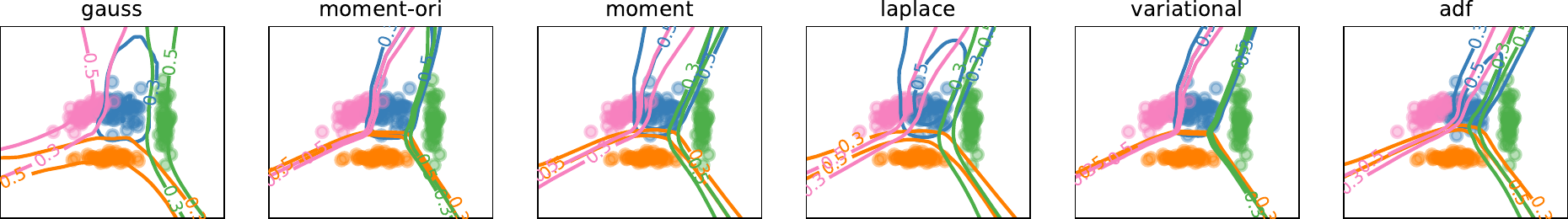}
    \caption{Performance of various methods including online approximate inference in the exact model [adf] and online exact inference in the approximate models [others] on a toy dataset.}
    \label{fig:fourclass_online}
\end{figure}

We first visualise the predictions on a toy dataset, made by various online inference strategies in \cref{fig:fourclass_online}

\begin{figure}[!ht]
    \centering
    \includegraphics[width=0.8\linewidth]{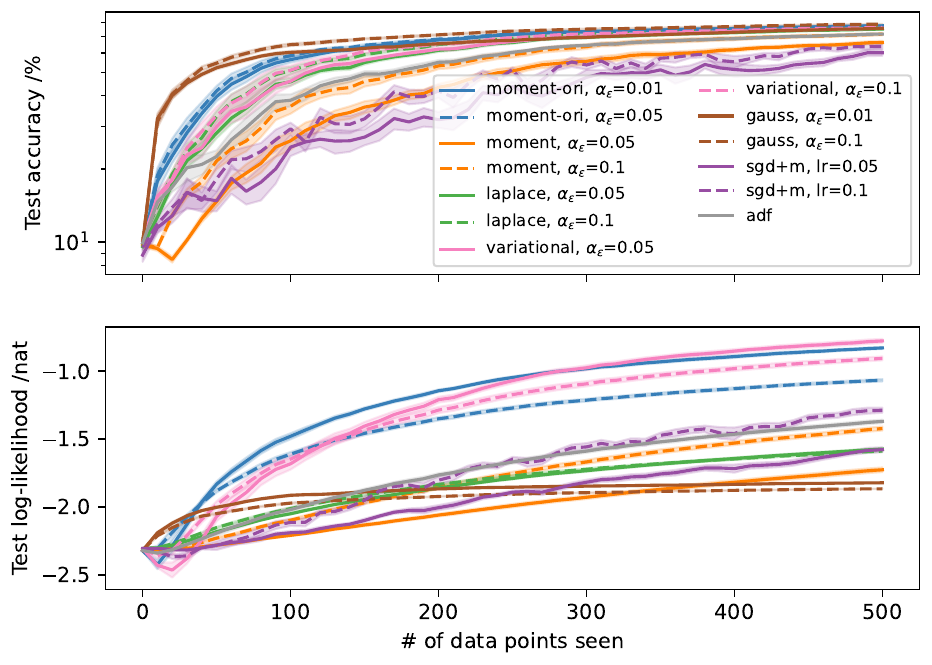}
    \caption{Online inference with streaming MNIST data using SGD, approximate inference (ADF) or exact inference in the approximate model (gauss, moment-ori, moment, laplace, variational).}
    \label{fig:mnist_kalman}
\end{figure}

In addition to the experiments on FashionMNIST (with random features) and CIFAR10 (with VGG-11 features) presented in the main text, we also ran an experiment on MNIST with random features and report the results in \cref{fig:mnist_kalman}.



\begin{figure}[!ht]
    \vspace{-12pt}
    \centering
    \includegraphics[width=0.6\linewidth]{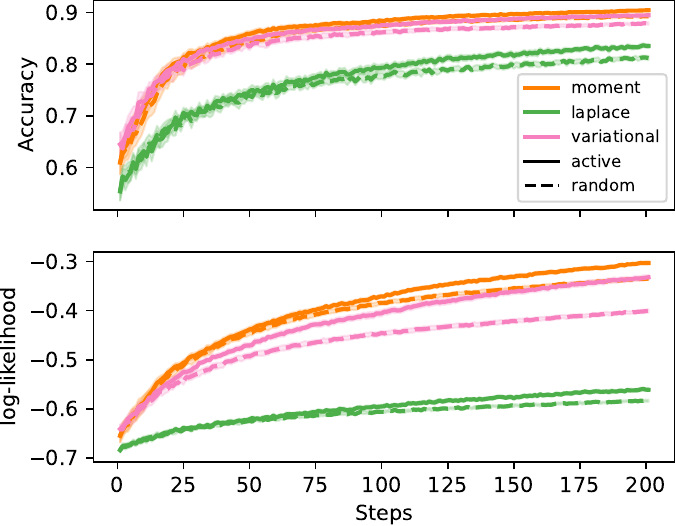}
    \vspace{-15pt}
    \caption{Active learning of a Bayesian logistic regression model on the IMDB dataset.}
    \label{fig:imdb_active}
    \vspace{-10pt}
\end{figure}

We visualise the performance after each active learning step for various likelihood approximations in \cref{fig:imdb_active}. The set up is described in the main text. We did not experiment $\alpha_\epsilon$ here and just kept them fixed at $0.1$.

\begin{figure}[!ht]
    \centering
    \includegraphics[width=\linewidth]{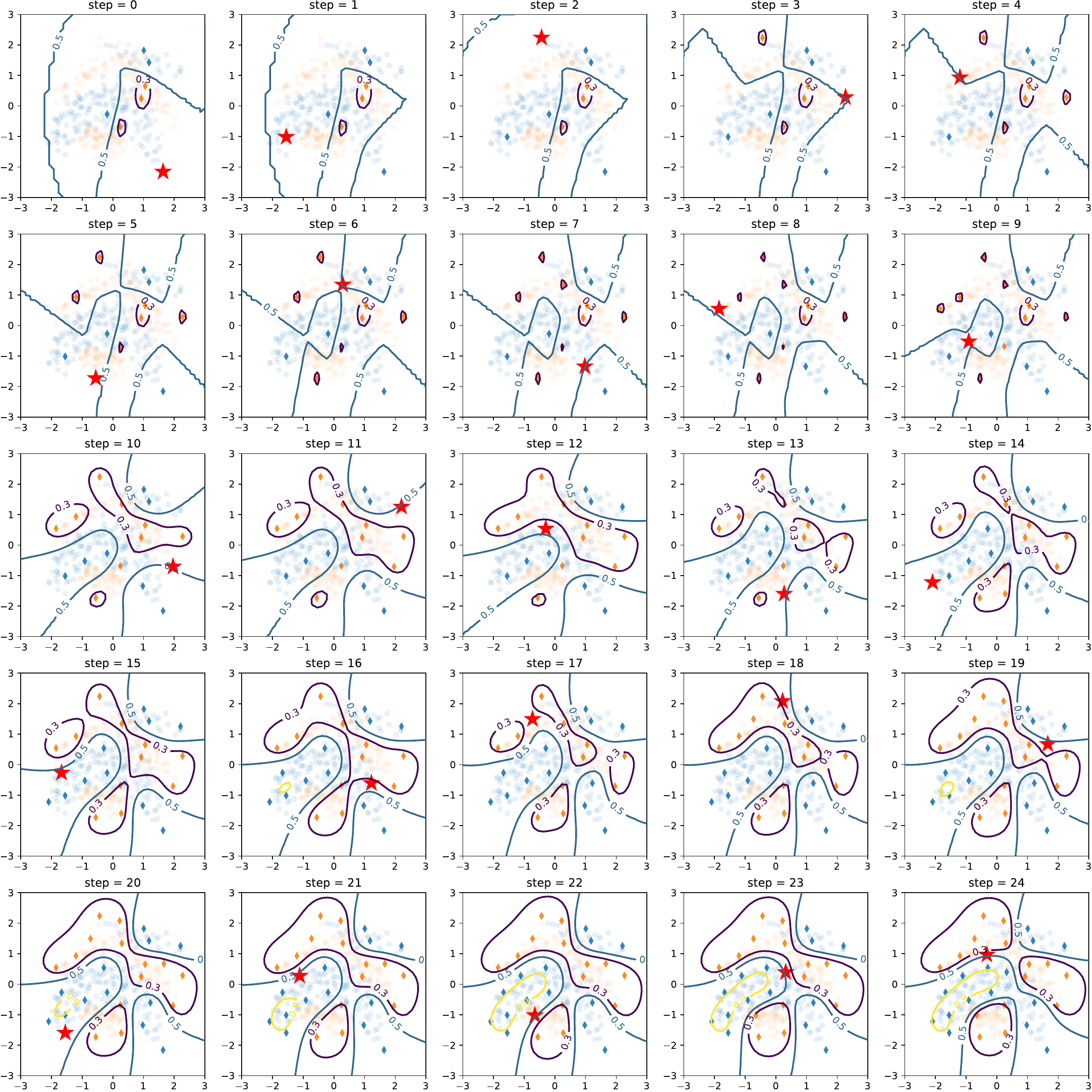}
    \caption{Active learning using variational Gaussian process classification.}
    \label{fig:active_gpc}
\end{figure}

\begin{figure}[!ht]
    \centering
    \includegraphics[width=\linewidth]{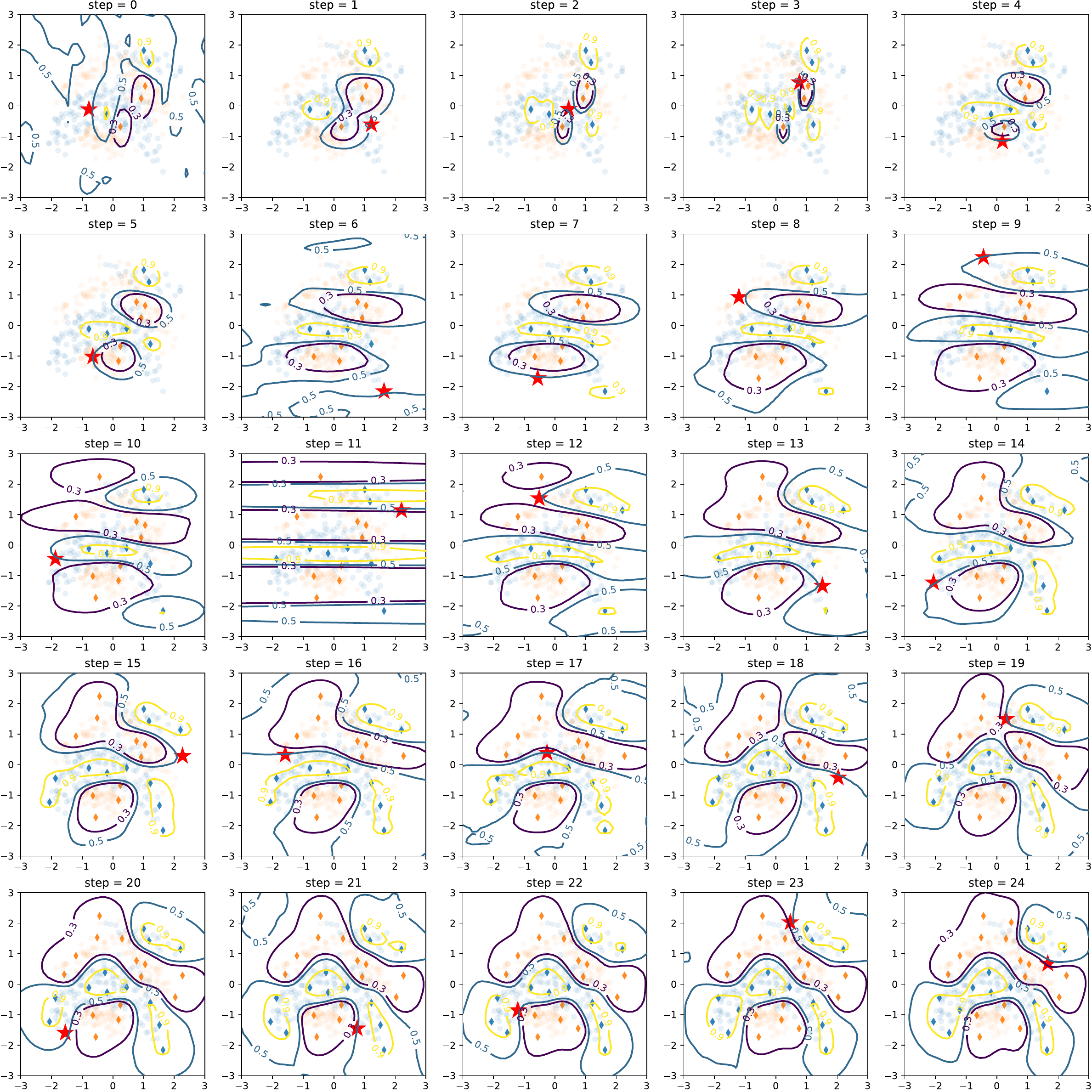}
    \caption{Active learning using Gaussian process regression with the variational matching likelihood approximation in \cref{sec:beta_binary}.}
    \label{fig:active_gpr}
\end{figure}

We also visualise in \cref{fig:active_gpc,fig:active_gpr} pool-based active learning steps on a toy classification dataset using variational inference in the exact GP binary classification model, or exact inference in the GP regression model with the variational likelihood approximation in \cref{sec:beta_binary}. 

\end{document}